# Developing a Multi-task Ensemble Geometric Deep Network for Supply Chain Sustainability and Risk Management

**Mehdi Khaleghi[a] , Nastaran Khaleghi[b] , Sobhan Sheykhivand[c], Sebelan Danishvar[d]**

[a] Faculty of Industrial Engineering, Azad University of Tehran, Tehran, Iran
[b]Faculty of Electrical and Computer Engineering, University of Zanjan, Zanjan; khaleghi.nstr@tabrizu.ac.ir
[c]Department of Biomedical Engineering, University of Bonab, Bonab 55517-61167, Iran; s.sheykhivand@tabrizu.ac.ir
[d]College of Engineering, Design and Physical Sciences, Brunel University London, Uxbridge UB8 3PH, UK

## Abstract:

The sustainability of supply chain plays a key role in achieving optimal performance in controlling the supply chain. The management of risks that occur in a supply chain is a fundamental problem for the purpose of developing the sustainability of the network and elevating the performance efficiency of the supply chain. The correct classification of products is another essential element in a sustainable supply chain. Acknowledging recent breakthroughs in the context of deep networks, several architectural options have been deployed to analyze supply chain datasets. A novel geometric deep network is used to propose an ensemble deep network. The proposed Chebyshev ensemble geometric network (Ch-EGN) is a hybrid convolutional and geometric deep learning. This network is proposed to leverage the information dependencies in supply chain to derive invisible states of samples in the database. The functionality of the proposed deep network is assessed on the two different databases. The SupplyGraph Dataset and DataCo are considered in this research. The prediction of delivery status of DataCo supply chain is done for risk administration. The product classification and edge classification are performed using the SupplyGraph database to enhance the sustainability of the supply network. An average accuracy of 98.95% is obtained for the ensemble network for risk management. The average accuracy of 100% and 98.07% are obtained for sustainable supply chain in terms of 5 product group classification and 4 product relation classification, respectively. The average accuracy of 92.37% is attained for 25 company relation classification. The results confirm an average improvement and efficiency of the proposed method compared to the state-of-the-art approaches.

## 1. Introduction

A compulsory stage of leading business is supply chain management. The interactivity of various representatives of suppliers, producers, marketers, purveyors and consumers plays a significant role in managing the supply chain. The regulations are made for the representatives with the aim of influencing the interactivity between each part for enhancing transportation and distribution services. Making a plan to optimize the delivery, setting an appropriate balance of the demand and supply, documentation and identification of suppliers to prepare goods and services, and managing the return of products are the essential tasks in supply chain management [1].

There are some risks associated with the supply chain, and they should be detected. The procedure of detection, investigation and administration of these risks are the actions for risk mitigation in a supply chain. Also, risk management can be considered as the process of recognizing and managing the threats to the validity of services. There are expansive types of supply chain risks threatening the companies and their service offer capacities. Some examples of these risks are the risks of different catastrophes, political and geographical vulnerabilities, malwares, digital attacks and supplier financial failures [2]. An understanding of risks is necessary for companies to manage the risks. It is an essential part of management to have an insight and manage risks. The COVID-19 epidemic disease is a tangible example of risks for supply chain [3, 4]. Researchers developed a framework to handle the threatening risks of supply chain during the epidemic [5]. A pragmatic risk administration method requires identification of familiar and obscure risks, construction of a risk management system and execution of some techniques in order to reduce risks [6]. Extending the suppliers and making connections with them play an essential role in successful techniques for risk management in the supply network [7]. The investment in modern technologies is required for improving the strength, sustainability and discernibility of a supply chain [8].

Deep learning has been utilized in supply chain studies in recent years. Supply chain management includes different stages of planning, sourcing, manufacturing, delivery and returns. Prediction of the future production demand is considered for the first stage of management. Identification of suppliers is an important task of the second stage. The transportation of products and delivering them to the customers are performed in the third step. The fourth stage is processing product returns and refunds from the customers to mitigate costs [9]. A deep learning approach can be used to optimize the outcome of each stage. Analyzing the supply chains is a crucial stage in solving different chain management issues. Analyzing includes product classification, product relation classification and identification of anomalies for risk management. A vital methodology for classification, optimization, prediction and object detection in different domains is machine learning strategies and deep learning models.

Taking into consideration the connections between product feature vectors, plant locations and allocated resources, it is possible to represent the supply chain as a graph. The graph representation is the compulsory prerequisite of the graph deep learning. It is the first stage of analyzing the supply chain utilizing the graph structural representation.

As a result of merging the graph theory and machine learning, several architectures have been introduced. Training of the merged graph-based neural networks is performed by using geometric deep algorithms. The training algorithms make it possible to implement the deep networks according to the graph-based data in an efficient way [10].

There are a few number of studies with a focus on employing graph structure of the time-related input supply chain dataset and the feature extraction phase is an important step in them. The extracted features of the time-based records for the products have been utilized to train the networks in most of the proposed methodologies. The two-stage framework consisting of feature derivation and categorization impose a computational load to the procedure. Also, the non-graph deep networks suffer from ignorance of the connections between characteristics and data records of the products. In order to tackle these limitations, an effective graph illustration of data samples is introduced and a graph-based geometric deep network is introduced and developed to realize the delivery status for risk management. Besides, the product type classification and the edge connection classification are performed for increasing the sustainability supply network management. It provides an opportunity to consider the correlation and connections between time-series characteristics of the products in the supply chain management procedure. Furthermore, the suggested network structure anticipates a fast convergence by removing the feature derivation phase and improving the accuracy performance.

The contributions presented in this article can be introduced as follows:

(i) It suggests a hybrid geometric network for extracting discriminative patterns and identifying different categories in a supply chain.

(ii) The proposed technique uses graph illustration of features recorded for products. The correlation between the characteristics of products is employed to construct the graph.

(iii) The characteristics related to the products are utilized directly as the graph nodes in the proposed method in order to remove the feature extraction phase and decrease the calculation load in training the network.

(iv) The proposed ensemble method classifies the delivery status of the products hence improves the performance of risk administration in supply chain .

(v) The proposed network architecture provides a framework for 5 different product classification, 4 edge connections in terms of same product groups and 25 edge relation classification in terms of same plants. Hence, it develops the sustainability of the network.

(vi) It uses a parallel network of Chebyshev-based graph convolution and 1-D convolutional layers for sustainability and risk management.

The other sections of this paper are organized as follows. Section 2 unveils contemporary approaches of supply chain management using deep learning. In Section 3, the characteristics of the DataCo and SupplyGraph databases are provided. Also, the mathematical basis of graph convolution, Chebyshev graph convolution and graph attention networks are explained in this section. Section 4 describes the attributes and the structure of the extended ensemble network for different purposes. Some modeling targets include delivery status prediction, product classification and product connection classification for risk management and sustainable supply chain. Section 5 contributes and extends the experimental results. The figures provided in this section substantiate the performance of the proposed ensemble geometric deep network with the state-of-the-art techniques. Section 6 is allocated to the conclusions.

## 2. Related works

The deep learning approach has the ability to analyze large numbers of data and extract learned features with trained parameters. Machine learning algorithms help to forecast time-based features and sales prediction. The selection of routes in a supply chain network can be done with machine learning approaches and deep learning algorithms to reduce transportation and delivery costs and plan an efficient route. Deep learning models and classification algorithms can be used for improving the model accuracy for prediction of demand, detection of anomaly, optimization of logistics and sustainability of supply chain [9]. Some studies applied deep learning in order to forecast real-time data related with demand rate of the production. Also, the objectives of deep learning have been detection of the abnormal data, implementation a predictive and sustainable maintenance plan, promotion of decision-making [11-13]. Also, Deep learning in supply chain management has been used to propose methods for supplier selection [14, 15].

Pereira et al. in [14] proposed an analysis for selection of suppliers. Their approach was designed corresponding to the CRITIC-GRA-3N machine learning approach proposed by Almeida et al. [16] in 2022. The method improved the selection of auto parts dealer in the city of Guaratingueta-SP. It was able to rank and select the suppliers in an efficient way. In the study by Ramjan Ali et al. [15], authors identified a list of supplier selection benchmark which applies many organizations. Random Forest classification method in conjunction with RF-related feature extraction method have been used in this study. The most critical criteria for supplier selection investigated in [15] as quality, material price, information sharing and on-time delivery.

The deep networks have been used for introducing novel structures in order to optimize the supplier selection [17]. In another study by Yazdani et al. [17] in 2021, the interval valued fuzzy neurosophic (IVFN) model has been extended for selection of suppliers for a dairy enterprise in Iran.

In some studies, a multi-phase approach based on deep learning has been proposed for allocating the orders in a supply chain [18, 19]. In the study by Shidpour et al. [18] in 2023, a multi-objective model has been improved for developing the model performance. The objectives in their study have been considered allocating customer orders and selecting the suppliers. The corporate social responsibility scores have been considered to acquire the ideal result for the model in [18]. The impact of deep learning-based transformation on decreasing the cost of transfers and transaction expenses in production management has been investigated by Li et al. [20]. The other examples of deep learning applications of this approach in supply chain management are deep modeling for transportation and conveyance issues during wars [21], technology acceptance deep model and diffusion of innovation theory for production management in supply chain [22].

Another applicable field of study for deep learning models is low-carbon methodologies for green supply chain administration [23]. In the study by Chun Fu et al. [23] in 2023, the impact of low-carbon activities in construction industry have been investigated. The framework in their study has been based on the exploration of structural modeling based on least squares. The analysis of data has been done with the use of partial least squares in structural equation modeling (PLS-SEM). The analysis of supply chain relationship in [23] helped to propose plans in order to diminish carbon dioxide pollution emissions. The study conducted a positive effect on the surrounding environment. In the study by Niu et al. [24] in 2024, location choices for enterprises and allocated centers for distribution have been involved in modeling. The proposed model in [24] reduced the cost with appropriate location choice for allocation centers and production plants. Furthermore, it increased the reliability and improved the accuracy.

In the study by Sirina et al. [25], methods for logistics management have been applied in one of the largest companies of the world named "Russian Railways". The aim of their study has been handling transport services. The proposed modeling of the transport system optimized the transport system. The study helped the company to control the logistics in supply chain for transportation facilities [25]. Other applications of deep learning are about controlling the inventory considering the accuracy of demand forecasting [26], fraud recognition and designing a scheme for demand achievement rate [27]. The objective of mitigating the demand disruptions has been studied using a supply and production twin network [27]. The twin network of production in [27] helped to optimize the inventory and income levels. Also, it helped to solve the disruptions in fast-moving consumer goods industry. Besides, the technique fulfilled consumer convenience efficiently.

The deep learning for risk management has been done in recent years for different objectives such as selection and segmentation of suppliers for risk prediction in supply chain in 2021 [28]. Estimation of supplier responsiveness and improving the resilience and strength of the supply chain [27] are other examples of deep model application in risk management. The detection of disruption, fraud and anomaly are other objectives of deep models [29]. In the study by Sebastian Villa in 2022 [29], the authors made an exploration about the bullwhip effect in a supply chain. Some circumstances of horizontal competition between retailors have been considered during the study. A mathematical model [29] has been developed in a competitive system while two behavioral exploration have been done to analyze the impact of supplier and customer characteristics on the decision of the retailers. The results of the study [29] demonstrate that the competing for demand does not have an effect on how retailers dilate their orders, whereas competing for supply influence the participants' ordering decisions. Furthermore, the order variability decreased by up to 50% by modification of the supplier's strategy for distribution. The retailers ignore the order cancellations of the customers according to this study [29].

Since our suggested technique in this article utilizes a fusion of graph theory and deep learning, we have a brief summary of the history of the graph neural networks. In 2005, analyzing the graph illustration of data has been investigated using a graph neural network (GNN) architecture proposed by M. Gori et al. [26]. Considering a mapping function in order to convert graph to a Euclidean space, an algorithm has been proposed by Scarselli et al. in 2009 to approximate the parameters of the graph network with supervised learning [27]. Bruna et al. contemplated spectral convolutional kernels for graphs in 2013 [28]. Two years later, incorporating a graph approximation algorithm by Bruna and Hennaf, the spectral network have been developed to create a novel network [29]. In 2016, ChebNet was suggested by M. Defferrard et al. [30], and in the same year, another article published by Kipf and Welling developed an innovative kind of graph convolutional network (GCN) kernels[31].

The graph neural networks have been utilized by researchers in some studies related with supply chain management in recent years. Abushaega et al. in 2025 [30] considered a graph learning in local centers of the supply chain network to optimize the global supply chain network. They used the concept of federated learning in order to improve the sustainability of the global network. The graph construction phase in their study has been prominently based on the logistics and delivery services. The considered dataset was the Supply Chain Data dataset and was segmented into three subcategories: the distributer, the manufacturer and the supplier. The dataset characteristics in [30] pertain to the acquired raw material, the time of delivery and the costs of the supplier. Yuemei Sun in 2025 [31] used the concept of graphsage network learning for risk prediction in financial transaction network. An ATM security configuration for unusual activity detection has been proposed in [32] by Kshirsagar et al. based on graph network learning. The input video files have been collected from UCF crime database and DCSASS dataset in [32]. Abuse, shoplifting, arrest, burglary, explosion, assault, fighting, robbery, road accidents, shooting, arson and stealing have been considered for investigation in this research [32]. Their approach demonstrates the reliability for real-time surveillance applications. Foroutan et al. in 2024 [33] performed graph learning for classification purpose and investigation of price fluctuation in crude oil markets. The dataset in their work [33] encompasses considerable markets and zones for a long period of about two decades. A commuting flow with the use of graph neural networks has been proposed by Nejadshamsi et al. in 2025 [34]. Their algorithm was context-aware and they used geographical characteristics in their modeling. The database in [34] is about real-life dataset from Montreal Island in which the metro transit system plays a significant role in everyday commutation. The MTL Traject 2017 and the Survey of Origin Destination 2018 datasets have been used in the mentioned study. The datasets have been labeled trajectory data and commuting flow databases. They have been transmitted from persons' cellphones and the locations of the purchasers have been registered via the MTL Trajet travel application. The manufacturing service capability prediction with graph learning has been the aim of the study by Yunqing Li et al. [35].

In this paper, we propose an ensemble geometric network for risk administration in a supply chain, improving the sustainability of supply chain and automatic prediction of the order status delivery. In the next section, we delineate the dataset characteristics and the mathematical foundation of Chebyshev graph convolution kernel and graph attention layer.

## 3. Materials and Methods

In this section, the details of the two databases used in this study are explicated. The DataCo and the SupplyGraph databases are used in this study. Besides, the mathematical basics of Chebyshev graph convolution kernel and graph attention network will be elucidated to understand how the graph layers work in a graph deep network.

### 3.1. Database setting:

Table 1 illustrates the details of the DataCo dataset. This dataset consists of 52 columns for 180000 transactions for DataCo global company. Three types of transation, days for shipment (scheduled), days for shipping, benefit per order, sales per customer, latitude, longitude, order item discount rate, order item discount, order item total and order profit per order are the characteristics considered for each data sample. The target labels for these data samples are the late delivery risk status.

The eight kinds of order status including complete, closed, pending, cancelled, pending_payment, processing, on_hold, suspected_fraud, payment_review can be predicted via an eight-category classification model. Besides, the four types of shipping mode considering standard, first class, second class and same day would be classified with the four-category classification model.

Table1 . DataCo specifications.

| DataCo | Feature | Format | Min | Max | Description |
|---|---|---|---|---|---|
| 1 | Type | Word | 0 | 2 | Kind of executed transaction |
| 2 | Real number of shipping days | Digit | 0 | 6 | Actual working days for shipping activities of the purchased product |
| 3 | Scheduled number for shipment days | Digit | 1 | 4 | Number of days for scheduled transportation of the purchased product |
| 4 | Gain per order | Numeral | -613.77 | 186.23 | Gaining per order |
| 5 | Sales per customer | Numeral | 27.04 | 399.98 | Total sales percustomer made per customer |
| 6 | Latitude | Numeral | 18 | 44 | Latitude according to storage lacation |
| 7 | Longitude | Numeral | -120 | -66 | Longitude according to storage location |
| 8 | Order item discount | Numeral | 0 | 99.99 | The discount value of order item |
| 9 | Order item discount rate | Numeral | 0 | 0.18 | Discount percentage value of the order item |
| 10 | Total order item | Numeral | 9.37 | 479.95 | Total amount per order |
| 11 | Order profit rate per order | Numeral | -613 | 153 | Profit rate per order |

Table 2. Target feature specifications.

| DataCo | Target Feature | Format | Min | Max | Target Description |
| --- | --- | --- | --- | --- | --- |
| 1 | Late delivery risk | Binary | 0 | 1 | 1 for late delivery , 0 for on-time delivery |

The SupplyGraph is the second dataset utilized in this research. It consists of the features about products, companies and resources. It consists of nodes and edges labeled for connections between products for same groups and same plants, separately.

Fluctuations and variations of delivery to distributor, factory issue, production and sales order for each day have been considered for this dataset. The nodes are available for product group, product sub-group and storage location. Edges are available for connections between plants, connection between product groups. The characteristics for time-based features are in terms of units and weights. For homogenous type of SupplyGraph database, there are 40 products and there are five different categories for products and 25 numbers of plants.

The edge indexes corresponding to different plants are available in this database. The connections according to different product categories are gathered in a file. The graph construction can be done corresponding to plants and product categories and subgroups.

Table 3 . SupplyGraph dataset characteristics.

| **DataCo** | **Feature** | **Duration** | **Number of Products** |
| --- | --- | --- | --- |
| 1 | Temporal Data - Delivery to distributor | 2023/08/09 - 2023/01/01 | 40 |
| 2 | Temporal Data - Factory Issue | 2023/08/09 - 2023/01/01 | 40 |
| 3 | Temporal Data - Production | 2023/08/09 - 2023/01/01 | 40 |
| 4 | Temporal Data - Sales order | 2023/08/09 - 2023/01/01 | 40 |

Table 4 . Node connections in Homogenous SupplyGraph

| DataCo | Edge Type | Nodes | Number of connections |
|---|---|---|---|
| 1 | Plant | Products | 1647 |
| 2 | Product group | Products | 188 |
| 3 | Product Sub-Group | Products | 52 |
| 4 | Storage Location | Products | 3046 |

Figure 1 illustrates the DataCo characteristic signals for all transactions in the company. Figure 2 shows the fluctuations of production sales order, delivery to distributor and factory issue in accordance with four products of the SupplyGraph.

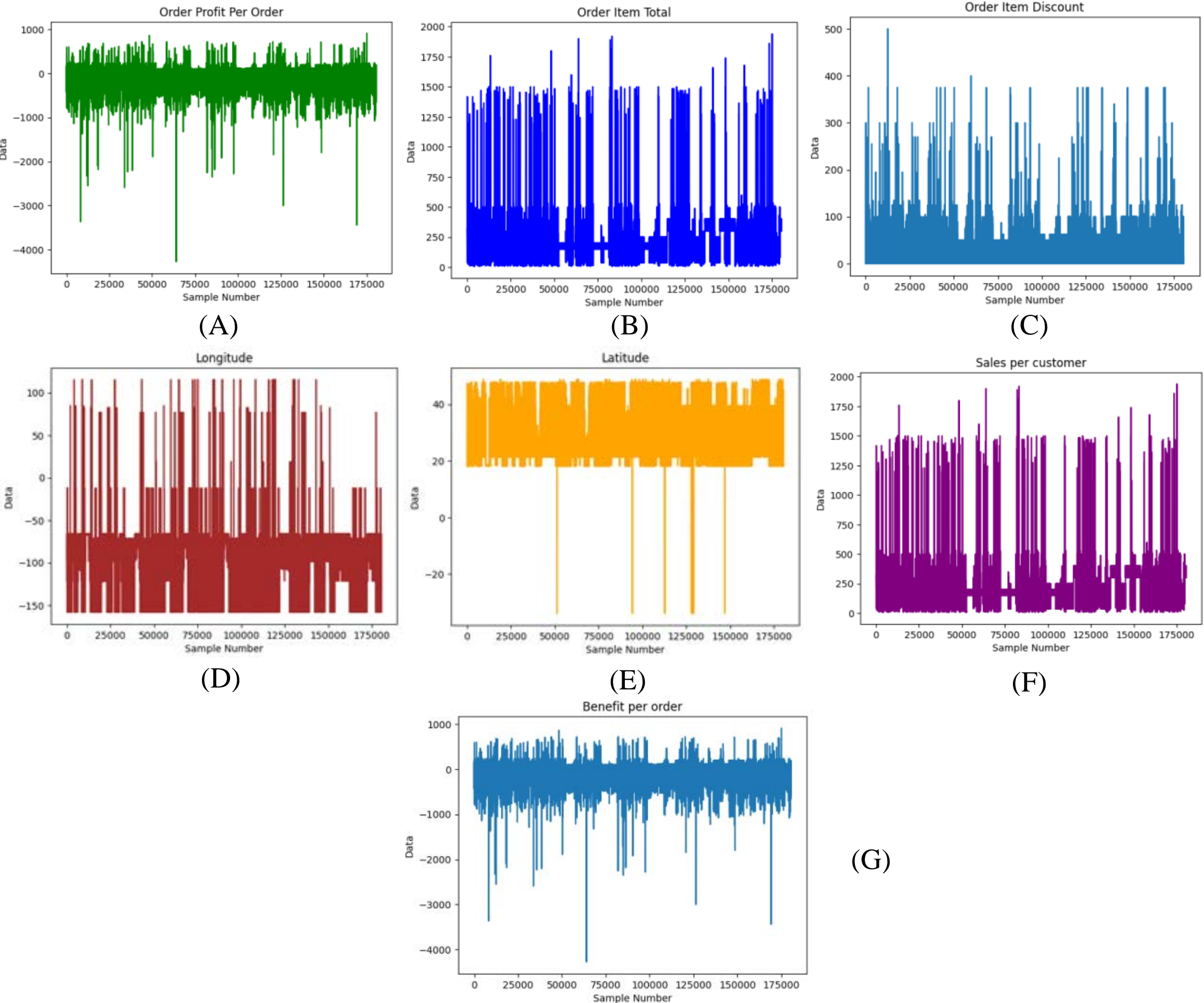

Figure1 . Characteristic plots for Dataco dataset. (A)Order profit (B) Order Item (C) Order discount (D)Longitude of location (E)Latitude of location (F) Sales (G) Benefit.

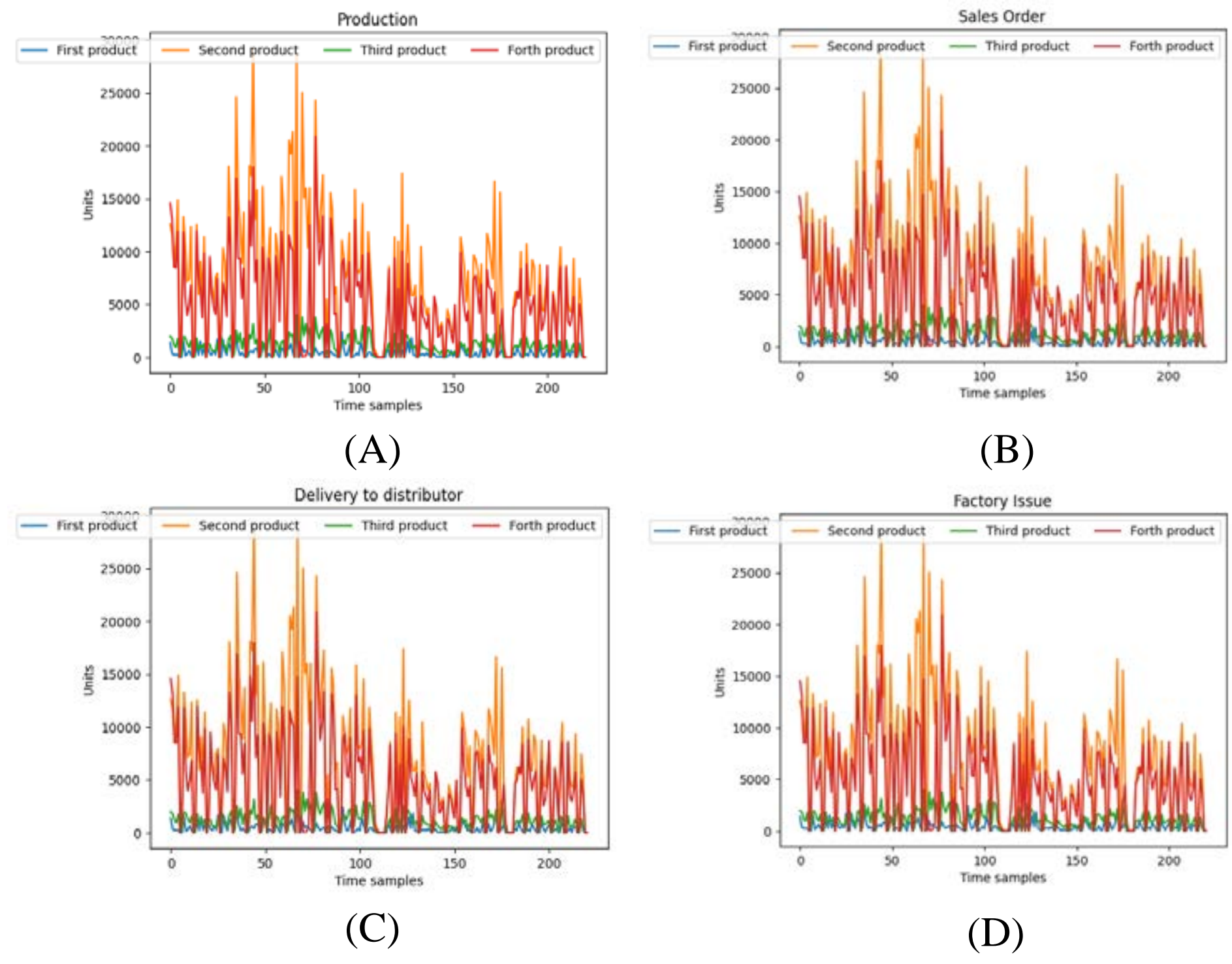

Figure 2 . Characteristic time plot for 4 different product ((A)Production,(B) Sales Order, (C)Delivery to distributor, (D)Factory issue)

### 3.2. Graph convolution

The research direction of Michaël Defferrard et al. [30] led to the popularization of graph signal processing (GSP). The mathematical functions in GSP take into account the attributes of the graph's elements and also the structure of the graph. GSP is utilized to develop the convolution kernels to the graph domain, and this area of research exploits signal processing techniques like the Fourier transform and deploys them to the graph representations. The use of Fourier transform in GSP results in graph spectral filtering, which is named graph convolution [32]. We explicate graph convolution layer in deep networks as described in [32]. Taking into consideration the graph structure, it is required to know the adjacency matrix and degree matrix according to the specific graph illustration. The $W \in \Re^{(N \times N)}$ is considered as the adjacency matrix and $D \in \Re^{(N \times N)}$ is corresponding to the degree matrix. The calculation of the ***i***-th diagonal component of the degree matrix can be described by (1).

$$D_{ii} = \sum_{j} w_{ij} \tag{1}$$

The Laplacian matrix of the graph named ***L*** in the formula is acquired by (2)

$$L = D - W \in \Re^{(N \times N)} \tag{2}$$

The fundamental operations in the graph domain are computed in accordance with the eigen vectors of the graph Laplacian matrix denoted by ***U***. These vectors can be obtained via the singular value decomposition (SVD) in (3).

$$L = U \Lambda U^T \tag{3}$$

The columns of $U = [u_0, ..., u_{N-1}] \in \Re^{(N \times N)}$ comprise the Fourier basis, and $\Lambda = diag([\lambda_0, ..., \lambda_{N-1}])$ is a diagonal matrix. Calculation of the eigenvectors returns the Fourier basis in accordance with the graph.

For a given signal $X \in \Re^N$ denoting the stacked feature vectors on the graph nodes, its graph Fourier transform (GFT) via the obtained graph basis functions is expressed as (4).

$$\hat{X} = (U^T) X \tag{4}$$

In formula (4), $\hat{X}$ designates the converted signal in the frequency domain and is the answer corresponding to the graph Fourier transform. The above formula expresses that the inverse of GFT can be obtained as the following form:

$$X = U(U^T) X = U\hat{X} \tag{5}$$

The filtered version of ***X*** by (***L***) can be written as (6).

$$Y = g(L) X \tag{6}$$

Using the following formulation in (7), it us obvious that the graph convolution of ***X*** with the vector of ***U***g(***Λ***) is symmetrical to the kernel operation of (6).

$$\begin{aligned} \mathbf{y} &= g(\mathbf{L})\mathbf{x} = \mathbf{U}g(\mathbf{\Lambda})\mathbf{U}^T\mathbf{x} = \mathbf{U}(g(\mathbf{\Lambda})).(\mathbf{U}^T\mathbf{x}) \\ &= \mathbf{U}(\mathbf{U}^T(\mathbf{U}g(\mathbf{\Lambda}))).(\mathbf{U}^T\mathbf{x}) = \mathbf{x} *_g (\mathbf{U}g(\mathbf{\Lambda})) \end{aligned} \tag{7}$$

The $g(\Lambda)$ in formula (7), is expressed as (8).

$$g(\mathbf{\Lambda}) = \begin{bmatrix} g(\boldsymbol{\lambda}_0) & \cdots & 0 \\ \vdots & \ddots & \vdots \\ 0 & \cdots & g(\boldsymbol{\lambda}_{N-1}) \end{bmatrix} \tag{7}$$

### 3.3. Chebyshev graph convolution

In this section, we explicate a particular type of graph convolution entitled Chebyshev graph convolution, while replacing g($\boldsymbol{L}$) in (8) with Chebyshev polynomial of $\boldsymbol{L}$. As we described earlier, the graph convolution of $\boldsymbol{X}$ with $\boldsymbol{U}$g($\boldsymbol{\Lambda}$) can be calculated as (8).

$$\mathbf{y} = g(\mathbf{L})\mathbf{x} = g(\mathbf{U}\boldsymbol{\Lambda}\mathbf{U}^T)\mathbf{x} = \mathbf{U}g(\boldsymbol{\Lambda})\mathbf{U}^T\mathbf{x} \tag{8}$$

The estimation of the g($\boldsymbol{\Lambda}$) is done via the $K$-order Chebyshev multinomials. The normalized version of $\boldsymbol{\Lambda}$ is utilized for approximation of the g($\boldsymbol{\Lambda}$) operation. The largest element among the diagonal entries of $\boldsymbol{\Lambda}$ is defined by $\lambda Max$ and the normalized $\boldsymbol{\Lambda}$ is computed with formula in (9).

$$\tilde{\boldsymbol{\Lambda}} = \frac{2\boldsymbol{\Lambda}}{\boldsymbol{\Lambda}_{\max}} - \mathbf{I}_N \tag{9}$$

where $\boldsymbol{I}_{\mathrm{N}}$ is the $\boldsymbol{N} \times \boldsymbol{N}$ identity matrix and the diagonal elements of $\tilde{\boldsymbol{\Lambda}}$ lie in the interval of [−1,1]. Approximation of ($\boldsymbol{\Lambda}$) based on the $K$-order Chebyshev polynomials framework is as formula (10).

$$g(\boldsymbol{\Lambda}) = \sum_{k=0}^{K-1} \boldsymbol{\theta}_k \mathbf{T}_k(\tilde{\boldsymbol{\Lambda}}) \tag{10}$$

In formula (10), $\boldsymbol{\theta}_{\mathrm{k}}$ denotes the coefficient of Chebyshev polynomials, and $\boldsymbol{T}_{\mathrm{k}}(\tilde{\boldsymbol{\Lambda}})$ can be acquired according to the following formulas in (11).

$$\begin{cases} \mathbf{T_0}(\tilde{\boldsymbol{\Lambda}}) = 1, \mathbf{T_1}(\tilde{\boldsymbol{\Lambda}}) = \tilde{\boldsymbol{\Lambda}} \\ \mathbf{T_k}(\tilde{\boldsymbol{\Lambda}}) = 2(\tilde{\boldsymbol{\Lambda}})(\mathbf{T_{k-1}}(\tilde{\boldsymbol{\Lambda}})) - \mathbf{T_{k-2}}(\tilde{\boldsymbol{\Lambda}}) \ , \ k \geq 2 \end{cases} \tag{11}$$

According to (12), the graph convolution kernel in (8) can be defined using (10) as illustrated in (12).

$$\begin{aligned} \mathbf{y} &= \mathbf{U}\, g(\boldsymbol{\Lambda})\ \mathbf{U}^T\mathbf{x} \\ &= \sum_{k=0}^{K-1} \mathbf{U} \begin{bmatrix} \theta_k \mathbf{T_k}(\tilde{\boldsymbol{\lambda}}_\mathbf{0}) & \cdots & 0 \\ \vdots & \ddots & \vdots \\ 0 & \cdots & \theta_k \mathbf{T_k}(\tilde{\boldsymbol{\lambda}}_\mathbf{N-1}) \end{bmatrix} \mathbf{U}^T\mathbf{x} \\ &= \sum_{k=0}^{K-1} \theta_k \mathbf{T_k}(\tilde{\mathbf{L}})\mathbf{x} \end{aligned} \tag{12}$$

In formula (12), $\tilde{\mathbf{L}} = 2\frac{\mathbf{L}}{\boldsymbol{\lambda}_{\max}} - \mathbf{I}_N$ is the normalized type of the Laplacian matrix.

The expression of Chebyshev graph convolution in (12) delineates that it is symmetrical to the exploitation of the convolutional results of x with each parts of the Chebyshev multinomial.

### 3.4. Graph Attention

Attentional graph networks point out the restrictions of convolutional graph neural networks by enhancing optimizable self-attention procedures that allocate differing significance to different neighbors. In this way [36, 37], the interactions are implicit and the formula is defined in (13).

$$\vec{h}_u = \phi(\vec{x}_u, \underset{v \in N_u}{\oplus} a(\vec{x}_u, \vec{x}_v)\psi(\vec{x}_v)) \tag{13}$$

In (13), the operator **a** is an optimizable self-attention procedure that calculates the significant coefficients in an implicit way.

$$\alpha_{uv} = a(\vec{x}_u, \vec{x}_v) \tag{14}$$

They are often softmax normalized in accordance with all neighbors. The operator $\oplus$ in (13) is the summation and the accumulation is an affine mixture of the neighborhood node features. This technique enables the model to concentrate on more significant links and dependencies, to improve the prediction performance. The negative side of the fact is the rising of the calculation cost and the incremental trend of complexity. In supply chain logical analysis, attentional graph networks are specifically useful for anomaly detection and risk administration. Besides, it is advantageous in cases that the capabilities of the model to rate essential connections and obtain more precise interpretations.

## 4. Proposed Method

The graphical diagram of different stages in accordance with the proposed method is represented in Figure 3. The Dataco and SupplyGraph datasets are used in this study. As it can be seen in this figure, after the pre-processing of the data and graph design stage, the acquired graph would be applied to tune the parameters of the proposed Chebyshev ensemble graph network (Ch-EGN) during the training stage. The network includes two distinct parts of deep networks. The graph-based section consists of four layers of Chebyshev convolution layers and the convolutional part includes two sequential non-graph convolutional layers. The loss function of the ensemble network is the weighted summation of the parallel graph-based part and convolutional part of the network. The training phase of the Ch-EGN is performed with K-fold cross-validation.

### 4.1. Pre-processing stage:

The Dataco and SupplyGraph signals are considered in this study. The conversions of text-like features in DataCo to integers are the first step of the pre-processing stage. The selection of features in order to clean the DataCo is another step. The clean array of features is used and imposed to the graph design phase.

The target for training the proposed Ch-EGN is considered as the zero-one conversion of the on-time delivery status and late delivery into zero and one, respectively. For the SupplyGraph database, the target is the product number for product classification and is plant number, sub-product group number for edge classification purpose.

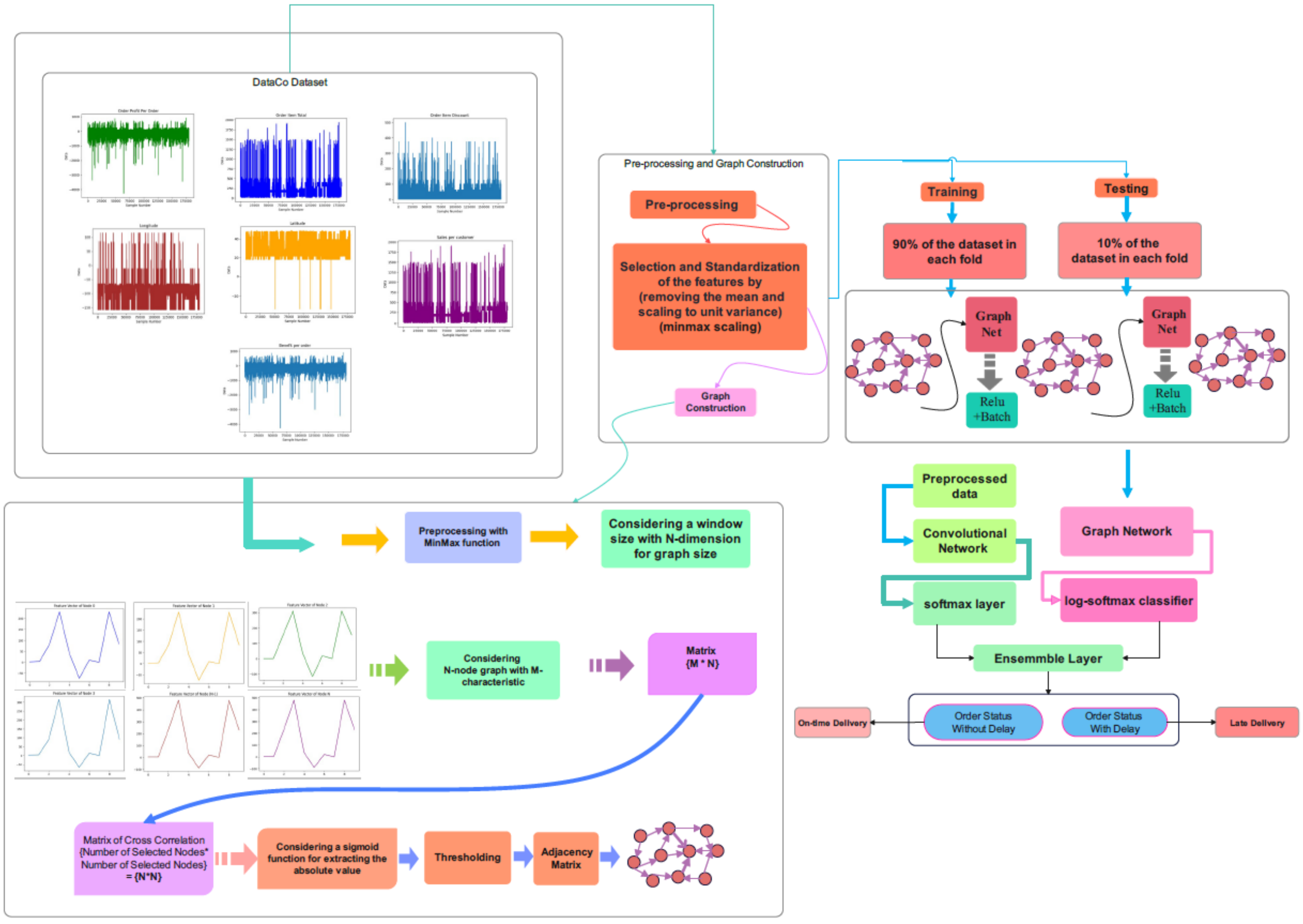

Figure 3 . The schematic overview of the proposed method.

### 4.2. Graph construction:

After pre-processing, the graph design stage is necessary to employ the acquired graph in the training phase of the proposed network architecture. The correlation of characteristic features in transaction data in DataCo is required for graph embedding. A sigmoid is utilized for computing the absolute value of the cross-correlation matrix. Also, a threshold level is considered in order to removing some non-zero elements of the output array. The adjacency matrix is the output of the sigmoid function and thresholding stage according to the simplified graphical representation of the graph design stage in Figure 2.

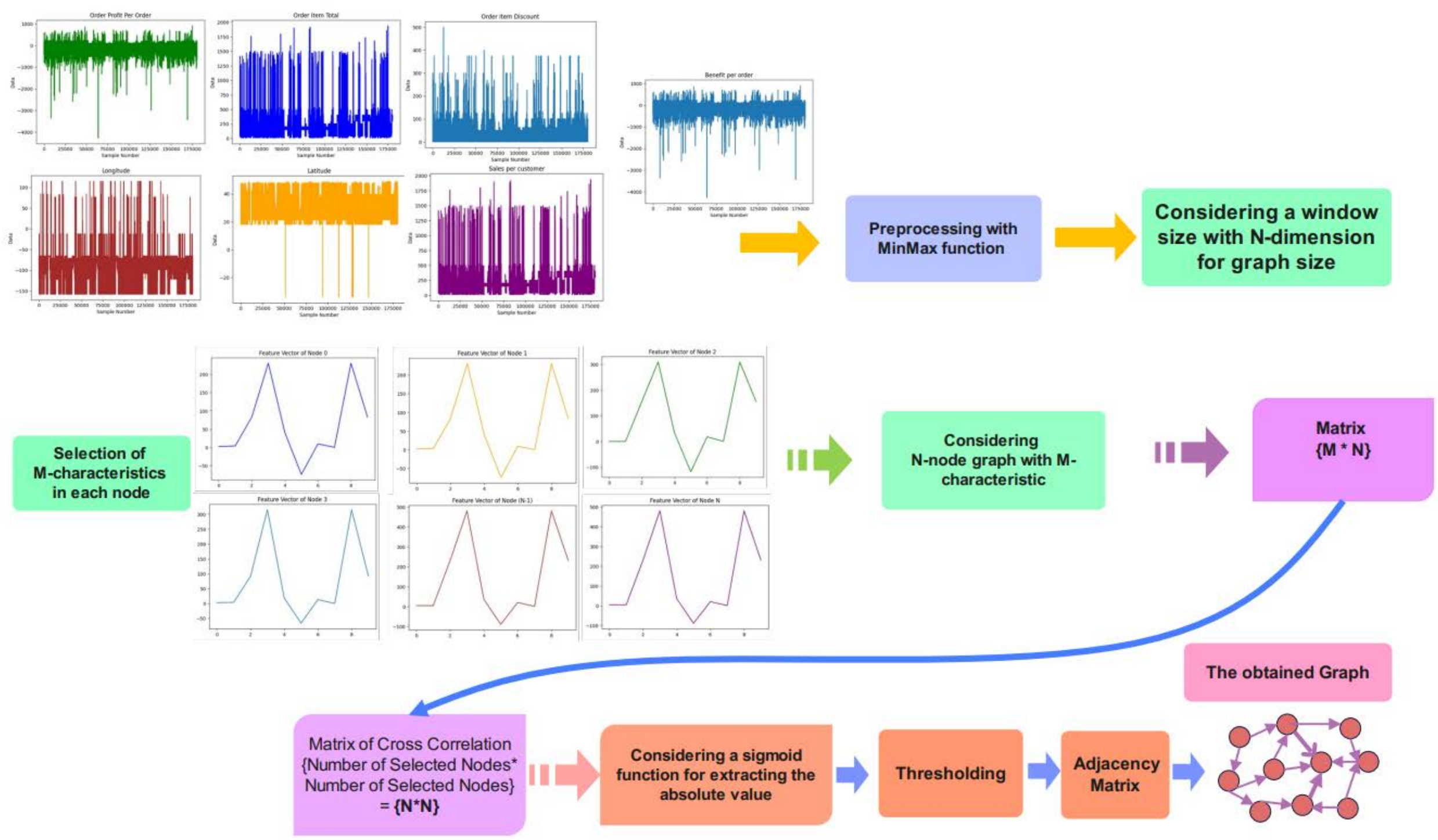

Figure 4 . Graph construction stage for DataCo.

### 4.3. Proposed Ch-EGN architecture:

Figure 4 delineates the detailed graphical representation of the proposed network architecture. As this figure shows, our proposed geometric Ch-EGN contains four layers of graph convolution. As specified by this figure, in every Chebyshev convolutional layer, the first step is the estimation of the Chebyshev convolution of the input graph via the graph Laplacian. The next layer is the activation layer. Also, a batch normalization is utilized in the output of each layer to normalize the input to the next layer.

The output of the pre-processing stage is imposed to the parallel convolutional part of the ensemble network. The loss function is the ensemble of two loss functions of the parallel parts of the Ch-EGN network. After log-softmax layers in parallel networks the obtained signal is classified according to the target vector.

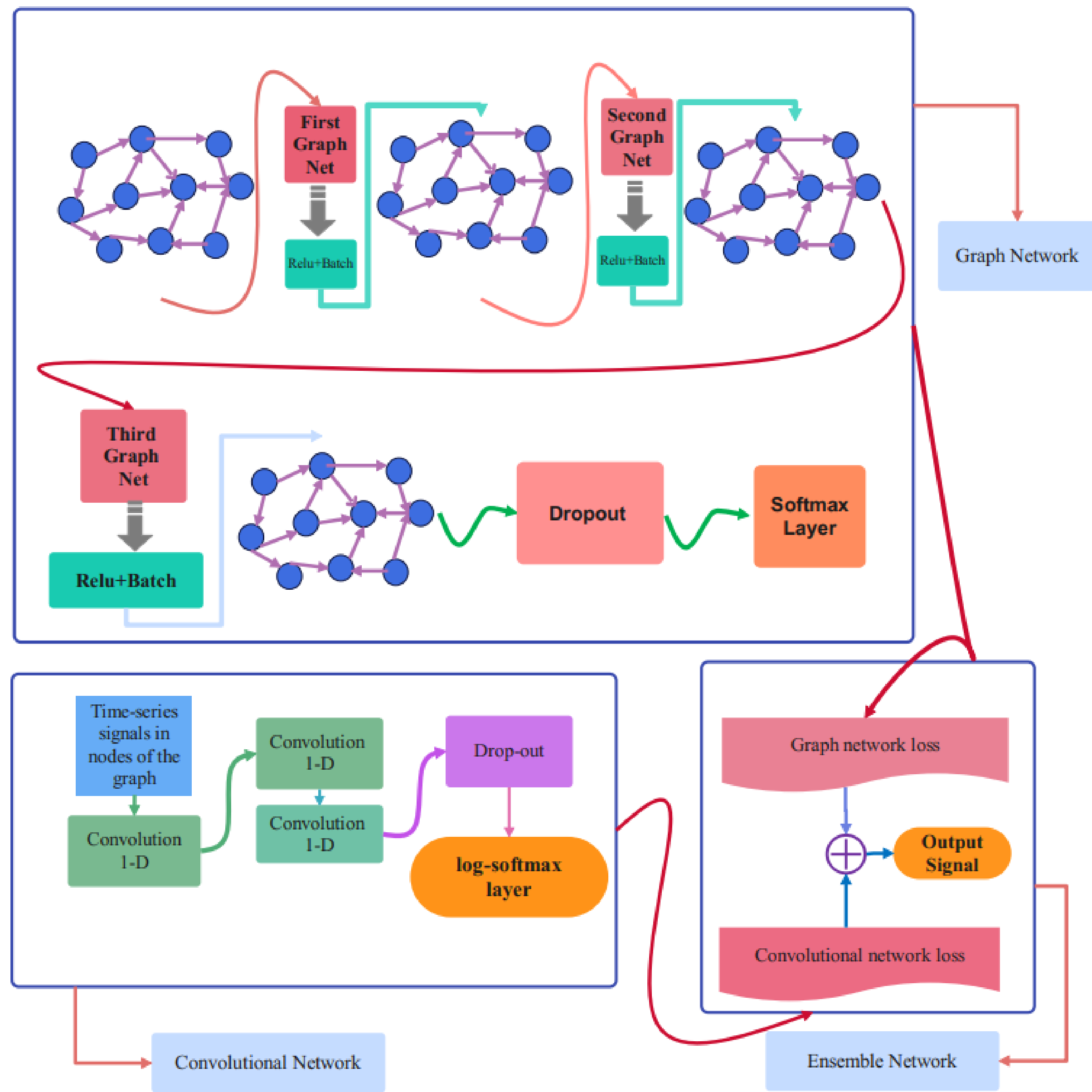

Figure 5. The detailed architecture of the proposed Ch-EGN.

Batch normalization makes the network to be stable throughout training procedure and the convergence of the network would happen more quickly. The normalization is allocated to each graph convolution layer. After four layers of Chebyshev convolution and two parallel convolution layers, the extracted feature array is acquired which is compatible with the size of target vector.

Table5 . Layers of the Graph section of the proposed method.

| **Layers** | | | **DataCo** | | | **SupplyGraph** | | |
|---|---|---|---|---|---|---|---|---|
| **Layer** | **Layer Name** | **Activation Function** | **Dimension of Weight Array** | **Dimension of Bias** | **Number of parameters** | **Dimension of Weight Array** | **Dimension of Bias** | **Number of parameters** |
| 1 | Chebyshev convolution layer | - | [1, 10, 10] | [10] | 110 | [1, 220,220] | [220] | 48,620 |
| 2 | Activation Layer | Relu | | | | | | |
| 3 | Batch normalization | - | [10] | [ 10] | 20 | [220] | [220] | 440 |
| 4 | Chebyshev convolution layer | - | [1,10,5] | [ 5] | 55 | [1, 220,100] | [100] | 22,100 |
| 5 | Activation Layer | Relu | | | | | | |
| 6 | Batch normalization | - | [5] | [ 5] | 10 | [100] | [100] | 200 |
| 7 | Chebyshev convolution layer | - | [1, 5, 2] | [2] | 12 | [1,100,20] | [20] | 2020 |
| 8 | Activation Layer | Relu | | | | | | |
| 9 | Batch normalization | - | [2] | [ 2] | 4 | [20] | [20] | 40 |
| 10 | Chebyshev convolution layer | - | [1, 2, 2] | [2] | 6 | [1,20,5] | [5] | 105 |
| 11 | Activation Layer | Relu | | | | | | |
| 12 | Batch normalization | - | [2] | [2] | 4 | [5] | [5] | 10 |

The details and characteristics of the proposed architecture are explicated in Table 5 and 6. Table 5 is related with the details of first part of the Ch-EGN. Table 6 is the attributes of layers matching to the convolutional part of the network. Also, it shows the kernel size for different layers, the size of strides in layers, number of kernels used for each layer and the total number of weights to be trained during the training procedure.

The target vector for delivery status prediction in DataCo is a two-class vector. The target vector for the SupplyGraph is 5 for product group classification, 4 for product sub-group classification and 25 for plant classification.

Table 7 demonstrates the weight parameters of the edge classification part of the network for classifying different categories of the edge connections of the graph.

Table 6 . Details of the convolutional part of the proposed method.

| Data | Layer | Layer Name | Activation Function | Output Dimension | Size of Kernel | Stride Shape | Number of Kernels | Number of Weights |
|---|---|---|---|---|---|---|---|---|
| DataCo | 1 | Convolution 1-D | LeakyReLU(alpha=0.1) | (10, 10, 5) | 1×5 | 1×1 | 10 | 510 |
| | 2 | Convolution 1-D | LeakyReLU(alpha=0.1) | (2, 10, 5) | 1×5 | 1×1 | 2 | 102 |
| Supply Graph | 3 | Convolution 1-D | LeakyReLU(alpha=0.1) | (100,220,5) | 1×5 | 1×1 | 100 | 110100 |
| | 4 | Convolution 1-D | LeakyReLU(alpha=0.1) | (5,100,5) | 1×5 | 1×1 | 5 | 2505 |

Table 7. Details of the edge classification part of the proposed method.

| Data | Layer | Layer Name | Activation Function | Output Dimension |
|---|---|---|---|---|
| SupplyGraph (product-based connections) (4 categories) | 1 | Linear | ReLU | (Number of edges, 100) |
| | 2 | Linear | ReLU | (Number of edges, 4) |
| SupplyGraph (plant-based connections) (25 categories) | 1 | Linear | ReLU | (Number of edges, 100) |
| | 2 | Linear | ReLU | (Number of edges, 25) |

### 4.4. Training and evaluation of the proposed Ch-EGN:

In training procedure, the generated input and target samples are utilized to tune the parameters of the suggested Ch-EGN to the Dataco and SupplyGraph datasets, we implement a 10-fold cross validation. After traing and tuning the variables and parameters of the Chebyshev graph convolution network and the parallel convolutional network, the testing phase is performed.

The training of the proposed Ch-EGN is performed according to the parameters in Table 5-7. The optimal weights are attained and summarized in this table. The cross-validation is selected for the validation procedure. The schematic view of this phase is indicated in Figure 6.

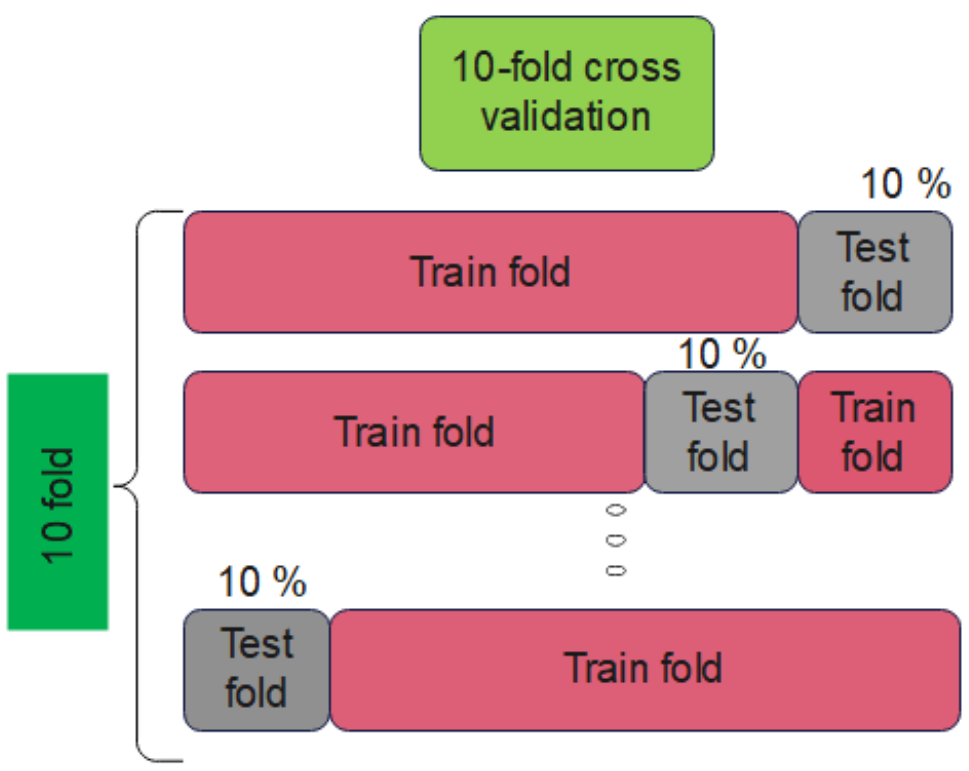

Figure6 . K-fold cross-validation stage.

A 10-fold cross-validation is fulfilled in accordance with Figure 6 using the training samples. The test stage can predict the delivery status of Dataco and classification purposes of SupplyGraph based on the calculated weights of the training stage. The pseudo-code in algorithm 1 explains the details of the proposed Ch-EGN. Table 6 shows the considered training search area and the optimal parameters for each scope.

Table 8. Details of training parameters.

| Parameters | Search Scope | Optimal Value |
|---|---|---|
| Optimizer of Graph section | Adam, SGD | Adam |
| Cost function of Graph segment | MSE, Cross-Entropy | Cross-Entropy |
| Number of Chebyshev convolutional layers | 2,3, 4 | 3 |
| Learning rate of Graph segment | 0.1, 0.01, 0.001 | 0.001 |
| Window size | 15,20,25,30 | 20 |
| Optimizer of convolutional segment | Adam, SGD | Adam |
| Learning rate of convolutional segment | 0.01, 0.001, 0.0001, 0.00001 | 0.0001 |
| Number of convolutional layers of second segment | 2, 3, 4 | 4 |

**Algorithm 1**

Chebyshev ensemble graph network (Ch-EGN)

**Input:** 1) Characteristic vectors $\boldsymbol{X}$, 2) A threshold level for adjacency matrix,

3) Chebysheev polynomial orders for each layer $K_1$, $K_2$, $K_3$, $K_4$,

4) Labeled train and test samples $\boldsymbol{Xtrain}$ and $\boldsymbol{Xtest}$,

**Output:** Class Labels for $X_{test}$

Initialize the model parameters.

Repeat according to the 10-fold cross-validation:

1: Determine the correlation co-efficient of the of $\boldsymbol{X}$ in $\boldsymbol{Xtrain}$.

2: Calculate the adjacency matrix $\boldsymbol{W}$ via using sigmoid function for the result of step 1.

3: Determination of the normalized Laplacian matrix $\hat{\Lambda}$.

4: Calculate the multinomials in accordance with the layer.

5: Extract the output of the four Chebyshev graph convolutional layer considering $K1$

and using $K2$ , $K3$ and $K4$ and the sequential activation layers.

6: Calculate the output of the dropout layer.

7: Calculate the output of the parallel simple convolutional layers.

8: Optimize the weights of the ensemble layers using appropriate loss function such as cross-entropy.

9: Update the weights of the layers using the totals ensemble cost function:

$$\begin{cases} Loss_{Cross-Entropy}(target, output_1) = -\frac{1}{n}\sum_{i=1}^{n}(target_i.\log output_{1i} + (output_{1i} - target_i).\log(target_i - real_i)) \\ Loss_{Cross-Entropy}(target, output_2) = -\frac{1}{n}\sum_{i=1}^{n}(target_i.\log output_{2i} + (output_{2i} - target_i).\log(target_i - real_i)) \end{cases}$$

$$Loss_{Total} = Loss_{Cross-Entropy}(target, output_1) + Loss_{Cross-Entropy}(target, output_2)$$

10: Attain the predictions for the embedded graphs in accordance with $\boldsymbol{Xtest}$ using the trained Ch-EGN.

Stop specifications: A maximum number of trials or acceptable accuracy.

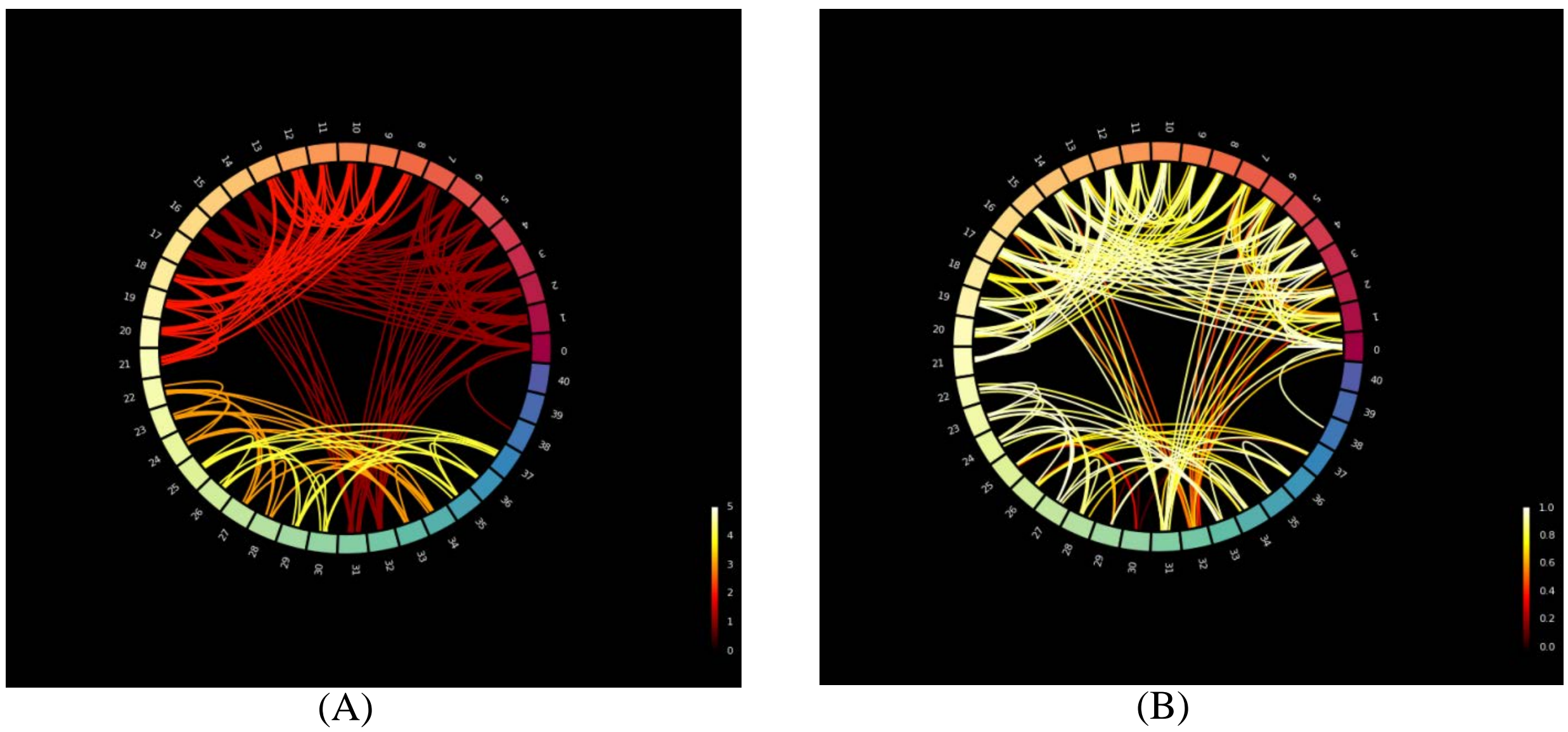

Figure 7. Circular connectivity between products in terms of product group edge indexes.
(A) Color Bar 1 (B)Color Bar 2.

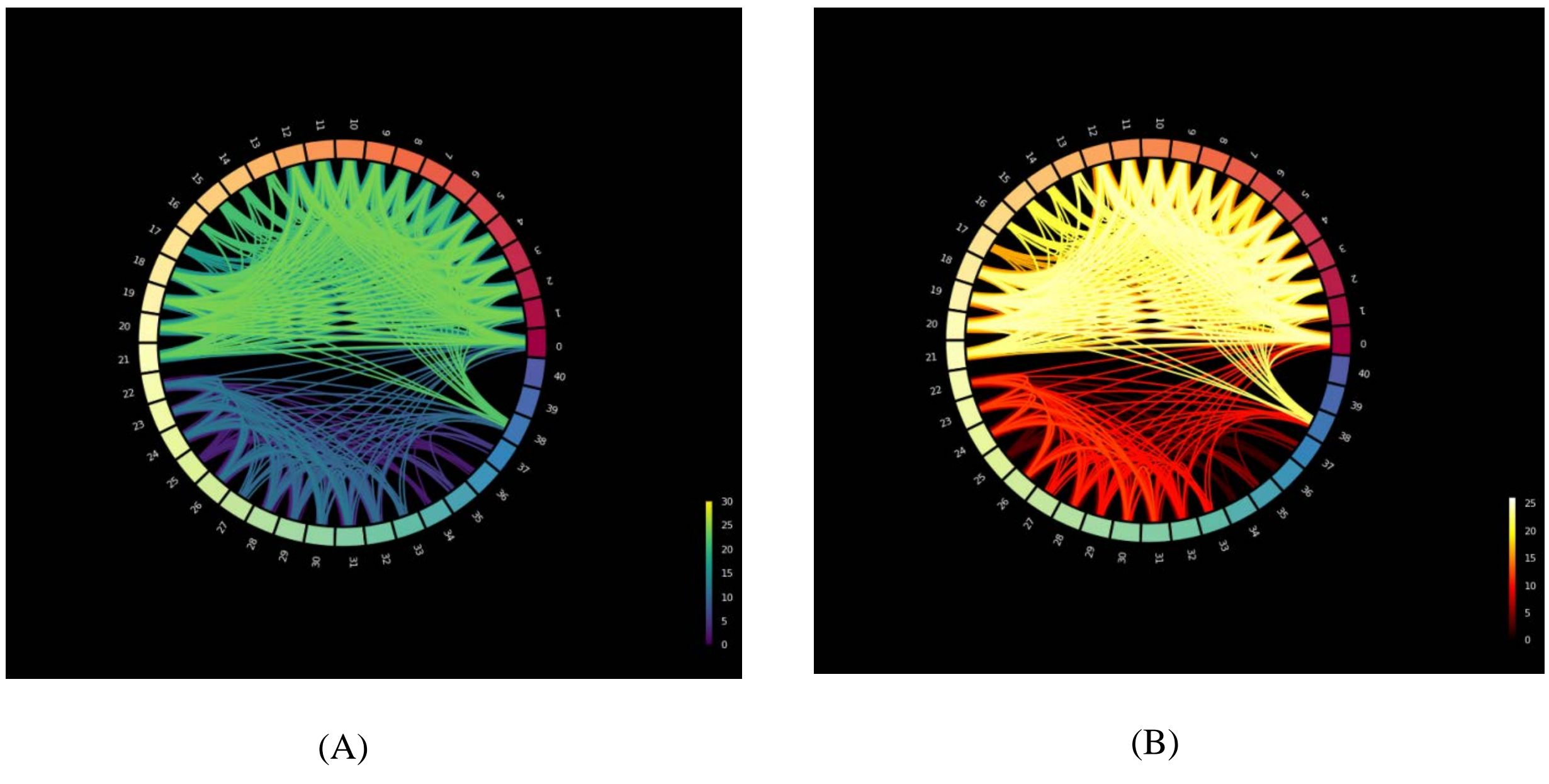

Figure 8. Circular connectivity between products in terms of plants edge indexes.
(A) Color Bar 1 (B)Color Bar 2.

Figure 7 illustrates the circular connectivity of edges in SupplyGraph dataset with similar product groups. The nodes are the products and the edge connections are the relation between products in terms of same product types.

Figure 8 is the circular connectivity between nodes of the SupplyGraph dataset considering the edge connections with similar plant locations.

There are 188 edge connections in Figure 7 and the connection are shown in two color maps to represent a good connection map. There are 1646 edge connectivity in Figure 8 in terms of similarity of the plant location.

## 5. Results and Discussion

In this section, the obtained results of analysis of the proposed Ch-EGN are presented. Our configuration is executed on a laptop with a 16 GB RAM, 2.8 GHz Core i7 CPU and a GeForce GTX 1050 GPU. The implementation of the proposed network is performed using the Google Colaboratory Pro platform.

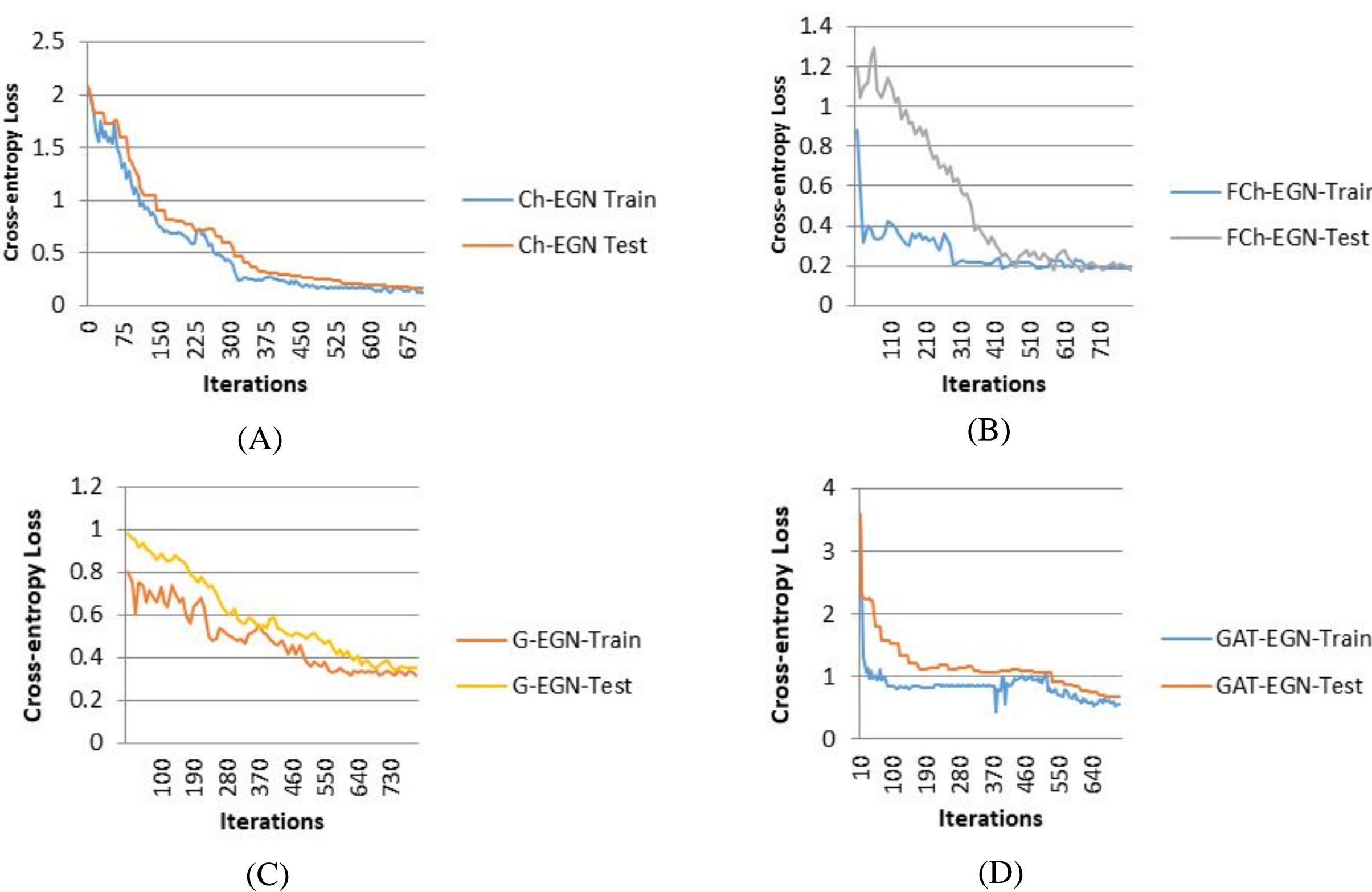

Figure 9 . Loss plots for training the DataCo. ((A) Ch-EGN (B) FCh-EGN (C) G-EGN (D) GAT-EGN)

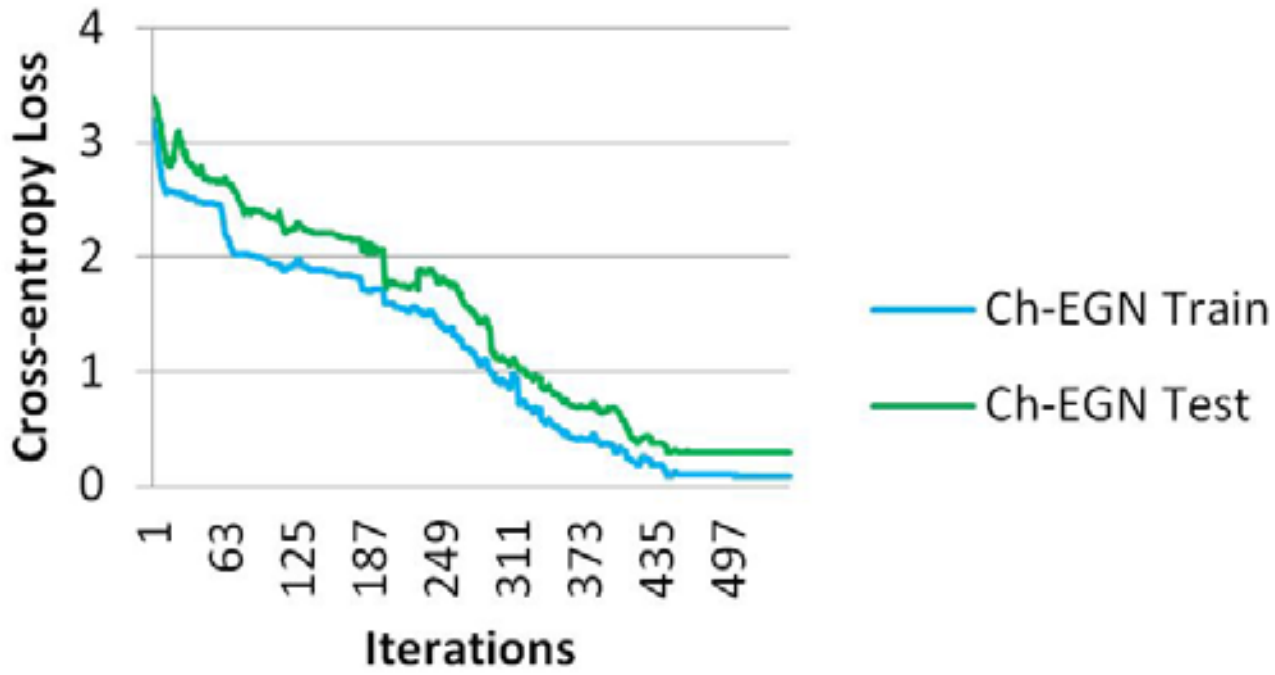

Figure 10. Loss plot for training the proposed method for the SupplyGraph.

(A) (B)

(C) (D)

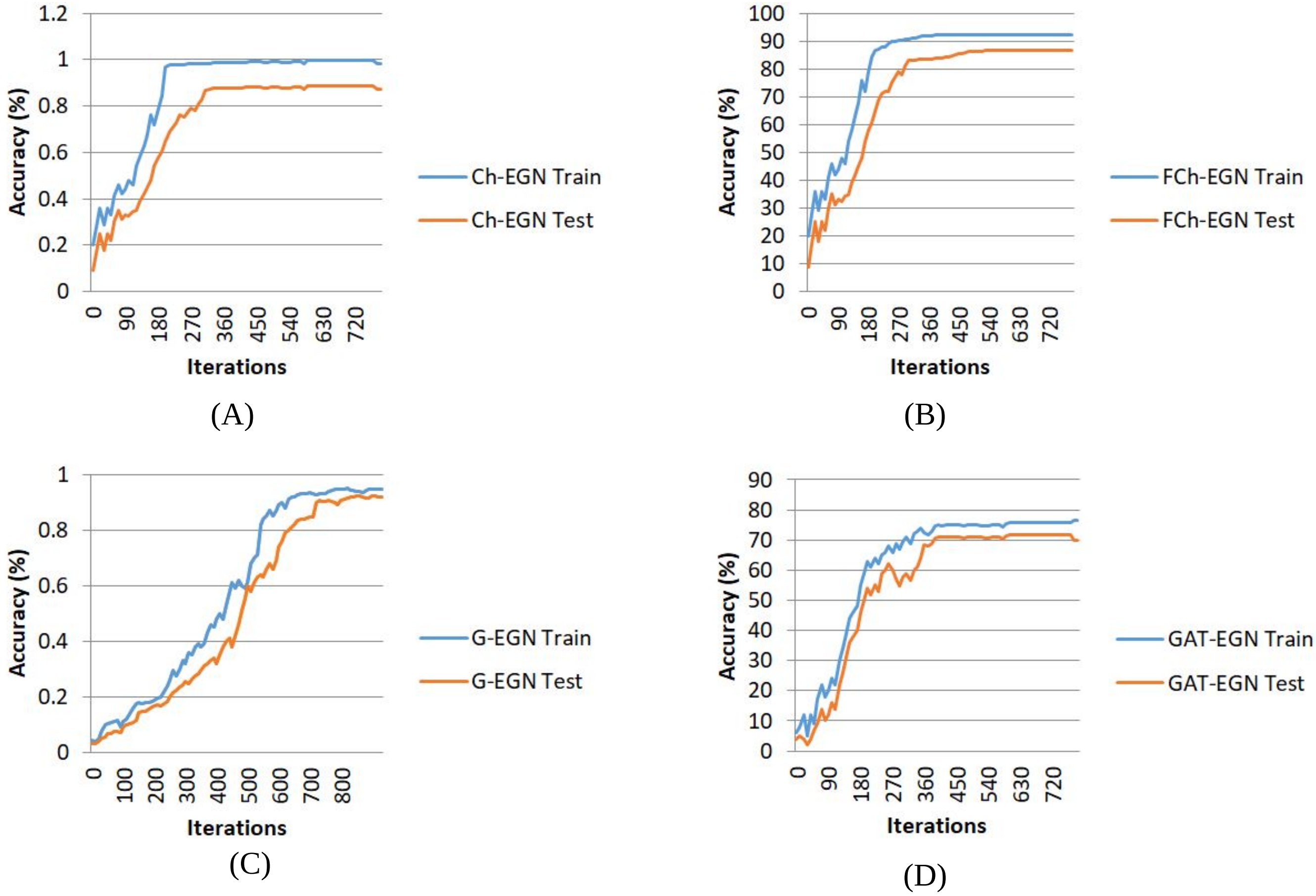

Figure 11. Accuracy plots for training the DataCo. ((A) Ch-EGN (B) FCh-EGN (C) G-EGN (D) GAT-EGN)

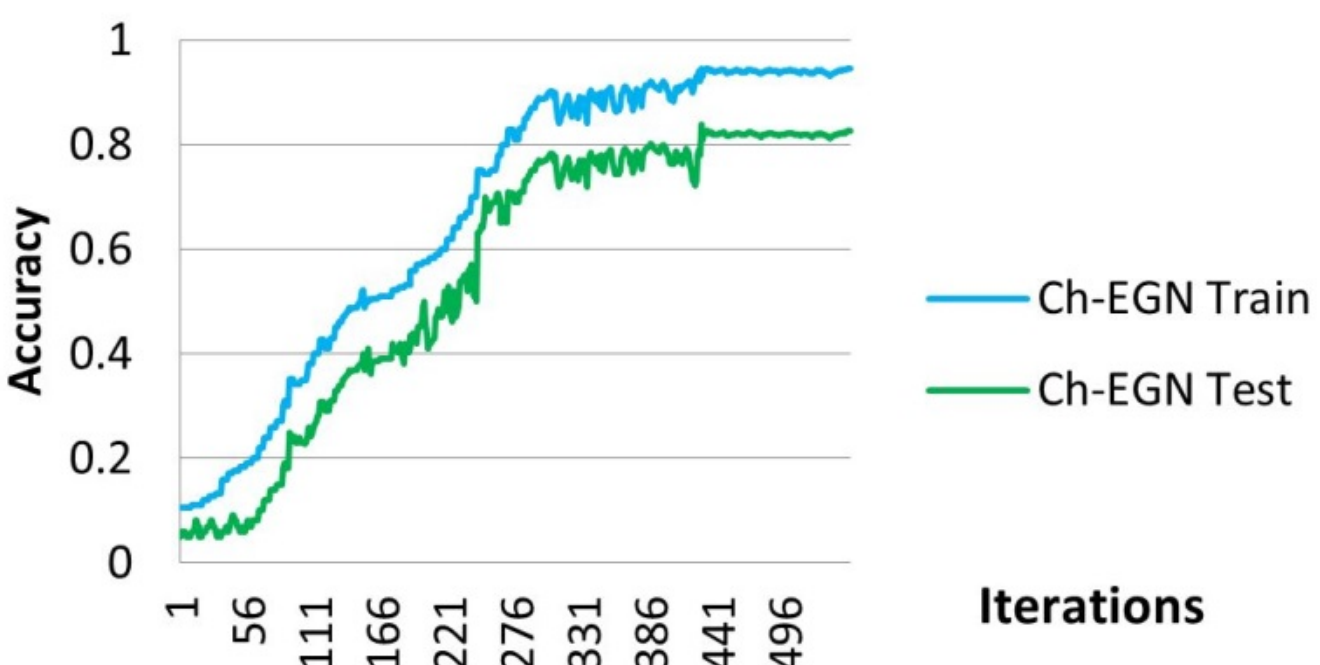

Figure 12. Accuracy plot for training the proposed method for the SupplyGraph.

Figures 9 and 11 show the performance of proposed Ch-EGN for DataCo based on the loss functions in accordance with the Chebyshev graph convolution segment and parallel convolutional network. Corresponding to this figure, Adam optimizer with optimal learning rate of 0.0001 and optimum weight decay of 4 × (10−4) taking into consideration the cross-entropy for the first segment of the network and the total loss corresponding to the pseudocode for the ensemble segment of the proposed network. This figure illustrates the loss plots for Ch-EGN, fuzzy Ch-EGN, G-EGN and GAT-EGN. The fuzzy needs more iterations in order to converge. The graph convolutional and graph attentional methods have a weak performance in comparison to the Chebyshev convolutional network. Three layers of graph convolution networks is considered for G-EGN and GAT includes three sequential layers of graph attention. As it can be seen, we consider more than 700 number of iterations for all methods considering a 10-fold cross-validation.

Figures 10 and 12 demonstrate the performance of the Ch-EGN for SupplyGraph. The number of repetitions necessary for the convergence of the proposed method with the aim of product type classification of this dataset is equal to 500. Considering the SupplyGraph dataset requires more than 500 iterations in order to convergence.

Table 9. Performance Metrics of the proposed method(Accurscy, Precision,Recall, F1-score)

| DataCo Category | Ch-EGN ($k_1$=1, $k_2$=1, $k_3$=1,$k_4$=1) | Ch-EGN ($k_1$=1, $k_2$=2, $k_3$=2, $k_4$=2) | Ch-EGN ($k_1$=2, $k_2$=2, $k_3$=2, $k_4$=2) | Ch-EGN ($k_1$=3, $k_2$=3, $k_3$=3, $k_4$=3) | FCh-EGN | GAT-EGN | G-EGN |
|---|---|---|---|---|---|---|---|
| on-time delivery | 98.2 | 98.3 | 94.7 | 93.2 | 95.3 | 93.8 | 91.8 |
| late delivery | 99.7 | 97.8 | 95.2 | 92.6 | 94.9 | 93.5 | 91.1 |
| **Overall accuracy** | 98.95 | 98.05 | 94.95 | 92.9 | 95.1 | 93.65 | 91.45 |
| **Precision** | 99.7 | 97.81 | 95.2 | 93.2 | 94.9 | 93.50 | 91.74 |
| **F1-score** | 98.9 | 98.04 | 94.7 | 92.87 | 95.09 | 93.64 | 91.41 |
| **Recall** | 98.22 | 98.32 | 94.72 | 93.15 | 95.28 | 93.78 | 91.09 |

Table 9 reports the performance metrics considering the DataCo for prediction of the delivery status for different methods. This table shows the on-time delivery and late delivery status prediction accuracy. Besides, it demonstrates the precision, F1-score and recall considering various orders for Chebyshev polynomial convolution layer and FCh-EGN, G-EGN and GAT-EGN methods.

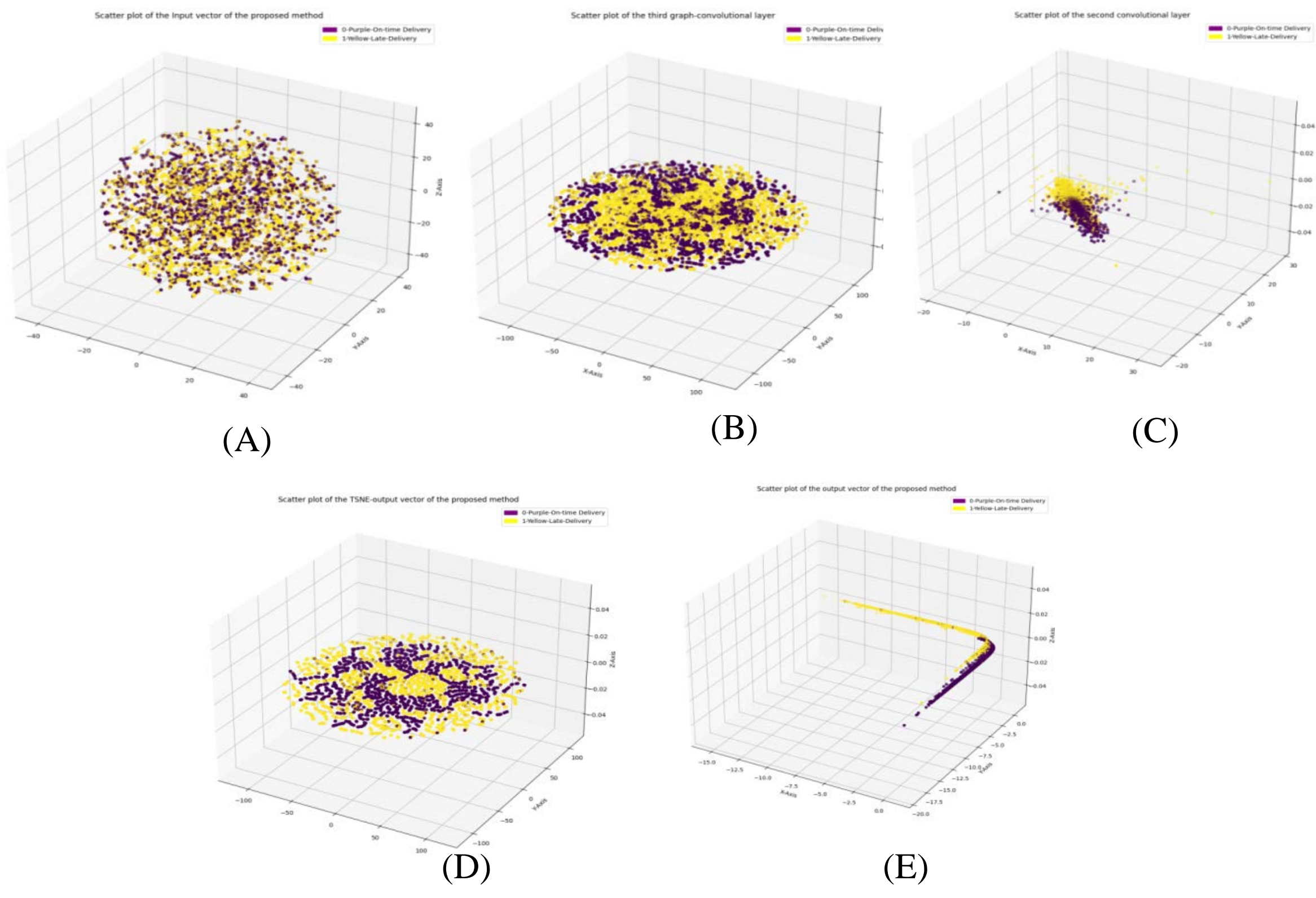

Figure 13. Three-dimensional TSNE plots for DataCo. ((A)input vector, (B) 3rd graph layer (C) 2nd convolution layer (D) Output layer (E) Output of the Softmax layer )

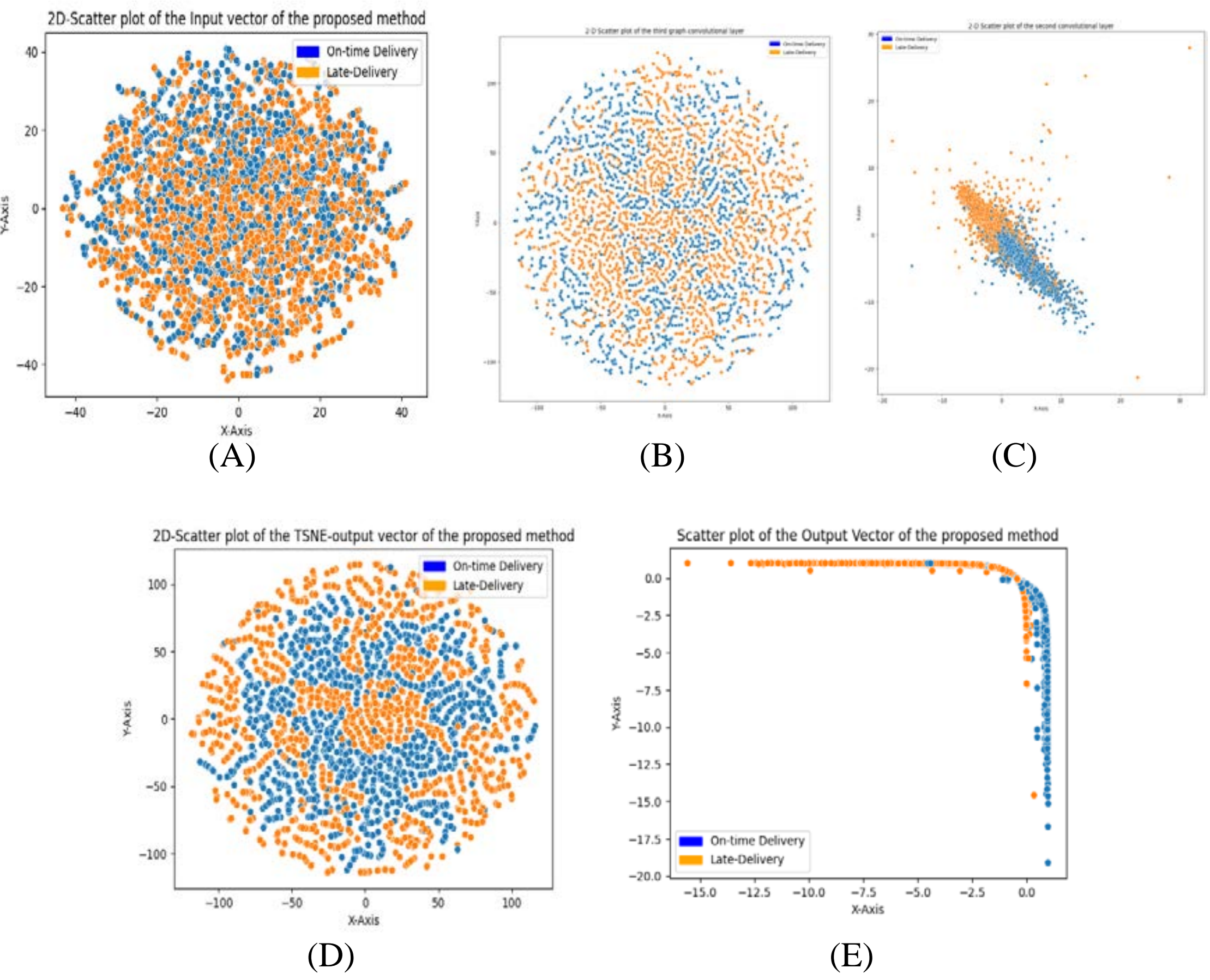

Figure 14. Two-dimensional TSNE plots for DataCo. ((A)input vector, (B) 3rd graph layer (C) 2nd convolution layer (D) Output layer (E) Output of the Softmax layer)

Figure 13 and 14 illustrate the three dimensional and two dimensional T-SNE plots for different layers of the proposed Ch-EGN in order to demonstrate the procedure of the classification and a tangible view of stages of classification considering the proposed Ch-EGN with DataCo dataset.

Table 10 is the report for the product classification in the case of considering SupplyGraph dataset. The classification accuracies for different types of product S, P, A, M and E can be seen in this table.

Table 10. Performance Metrics of the proposed method(Accurscy, Precision,Recall, F1-score)

| Supply Graph Categories | Ch-EGN | FCh-EGN | G-EGN | GAT-EGN |
|---|---|---|---|---|
| S | 100 | 90.09 | 85.75 | 80.6 |
| P | 100 | 85.28 | 82.92 | 78.9 |
| A | 100 | 87.39 | 82.25 | 80.1 |
| M | 100 | 83.27 | 81.29 | 75.8 |
| E | 100 | 84.46 | 81.95 | 76.3 |
| Overall accuracy | 100 | 86.54 | 84.57 | 78.9 |
| Precision | 100 | 86.08 | 84.18 | 78.4 |
| F1-score | 100 | 86.03 | 84.11 | 78.3 |
| Recall | 100 | 86.09 | 84.2 | 78.5 |

The results of the proposed Ch-EGN together with G-EGN, FCh-EGN and GAT-EGN are available in Table 10 for the SupplyGraph dataset. Moreover, this table presents the category-specific evaluation metrics thoroughly for all of these networks. As it can be seen in Table 10, the proposed Ch-EGN surpasses the other methods.

Table 11 shows the detailed results for classification of the product type, product relation classification in terms of products and product connection classification in terms of plant similarity. There are 5 different categories of product types. There are 4 different groups of relation based on the product relation and 25 different groups of connections for similar plants. This table confirms the good performance of the proposed method for node and edge classification in comparison to other methods. The proposed Ch-EGN outperforms other methods for classification of the nodes and edges of the SupplyGraph database.

Table 11. Accuracy for product relation classification.

| **SupplyGraph Dataset** | **Ch-EGN ($k_1$=1, $k_2$=1, $k_3$=1,$k_4$=1)** | **Ch-EGN ($k_1$=1, $k_2$=2, $k_3$=2, $k_4$=2)** | **Ch-EGN ($k_1$=2, $k_2$=2, $k_3$=2, $k_4$=2)** | **Ch-EGN ($k_1$=3, $k_2$=3, $k_3$=3, $k_4$=3)** | **FCh-EGN** | **G-EGN** | **GAT-EGN** |
|---|---|---|---|---|---|---|---|
| **Product group classification** | 100 | 100 | 100 | 100 | 86.54 | 84.57 | 80.09 |
| **Product group relation classification** | 98.07 | 98.07 | 96.1 | 94.47 | 85.2 | 81.54 | 79.32 |
| **Plant relation classification** | 92.37 | 92.37 | 90.03 | 88.68 | 82.3 | 80.76 | 78.44 |

Table 12 shows the performance metrics of the proposed method in comparison to the other novel and traditional methods. As it can be seen, our proposed geometric ensemble network outperforms the other conventional methods.

Table 12. Comparison with the conventional methods.

| **Method** | **Product group classification** | **Product group relation classification** | **Plant relation classification** |
|---|---|---|---|
| Ch-EGN | 100 | 98.07 | 92.37 |
| GNN-based [38] | 75.68 | 91.36 | 91.45 |
| KNN [39] | 64.44 | 74.75 | 74.63 |
| XGB [40] | 65.56 | 71.78 | 71.23 |
| Lgistic regression | 66.67 | 62.73 | 68.63 |

The confusion matrix is valuable way of confirming the efficiency of the proposed method Figure 15 shows the performance of the proposed Ch-EGN considering the DataCo dataset.

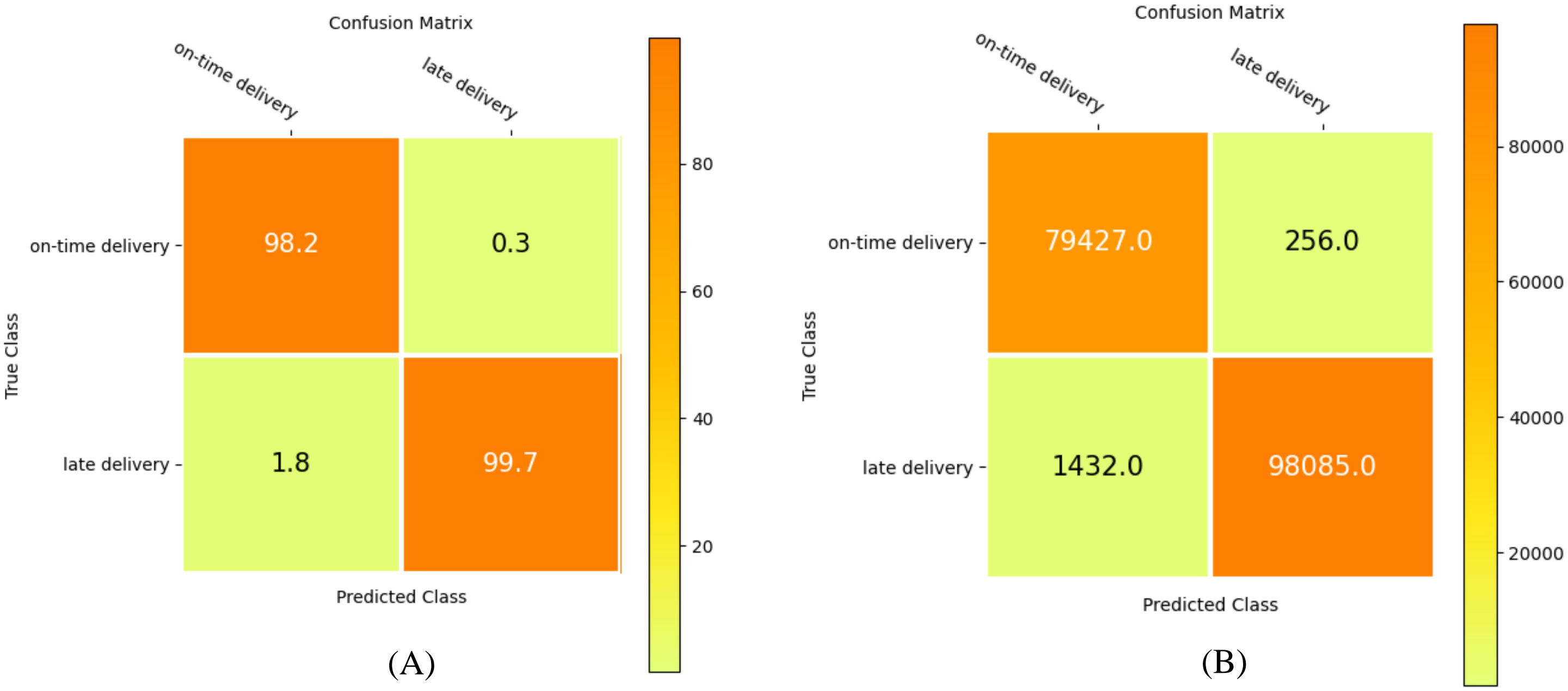

Figure 15. The confusion matrix for delivery status prediction. (A) Percentage (B) Number.

Figure 16 is the confusion matrix for the classification of the product types considering our proposed Ch-EGN.

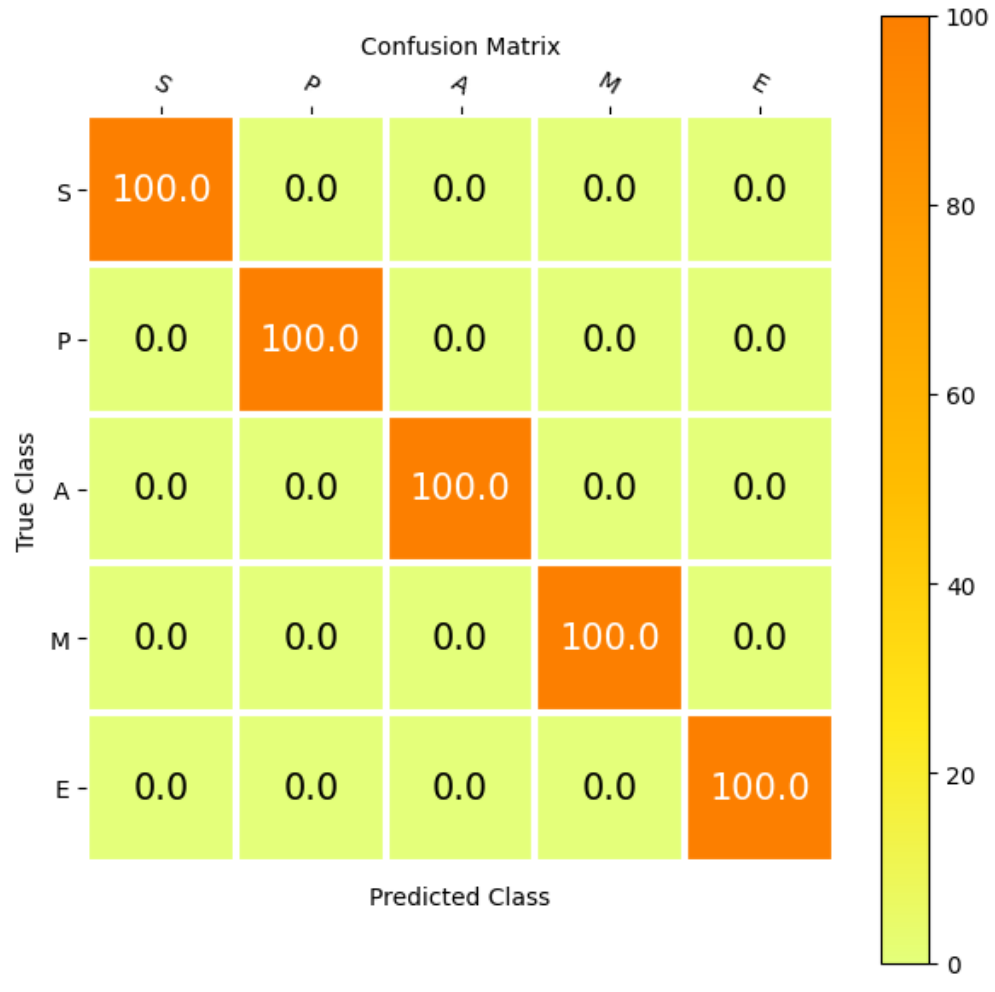

Figure 16. Product Classification.

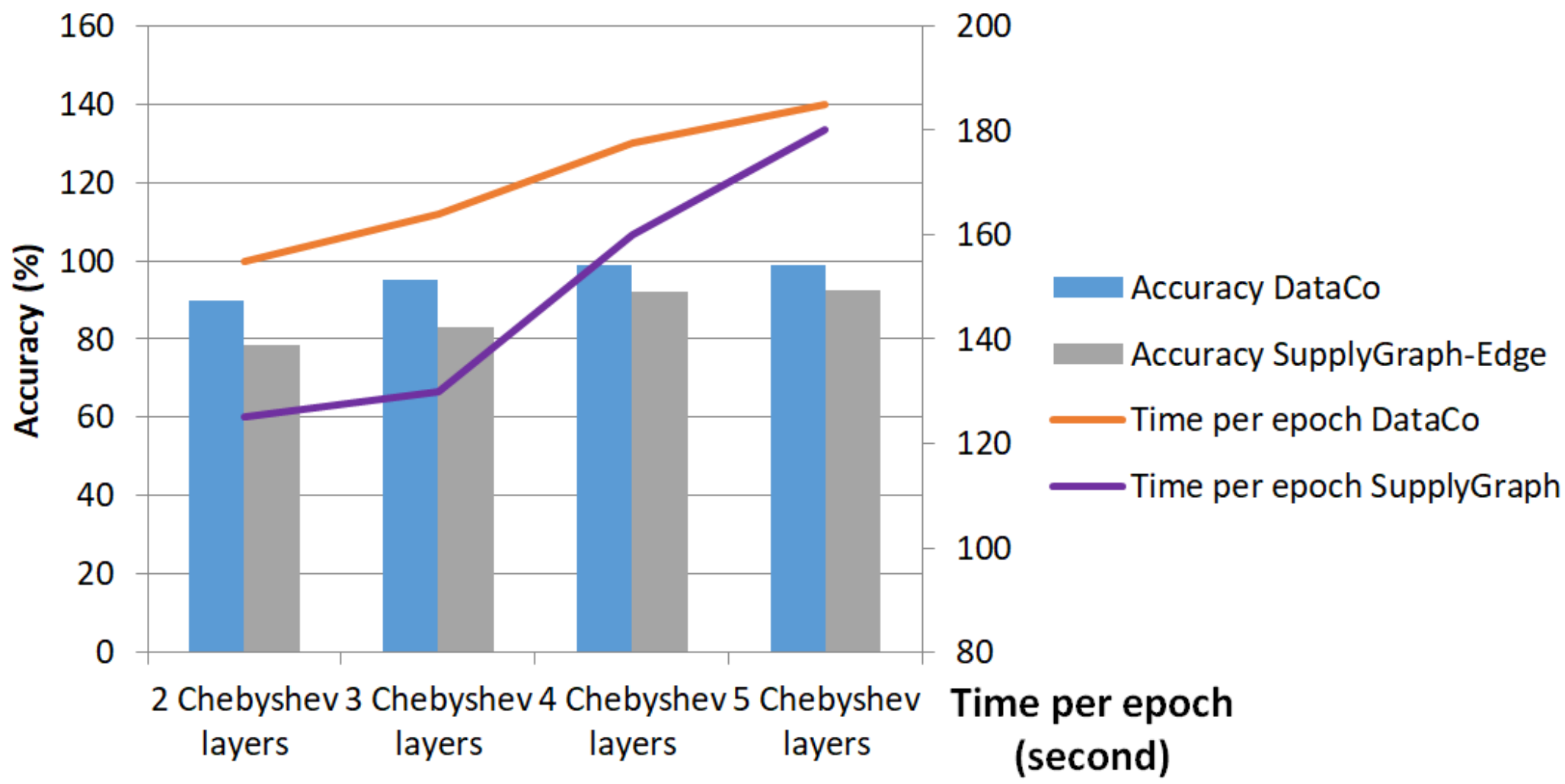

Figure 17. F1-score and time of training per epoch with different number of graph convolution layers.

To investigate the effect of different parameters on the optimality of the performance, we execute an extended experiment. In order to evaluate the effect of alternating the number of sequential Chebyshev filters, a series of training procedures are done for different numbers of sequential Chebyshev multinomial. Figure 17 showcases the results of tuning for 2, 3, 4 and 5 sequential Chebyshev layers. Setting the sequential layers more than four in this case study does not improve the performance, it affect the computational complexity. This figure showcases the incremental direction of the training time per iteration epoch of the proposed Ch-EGN.

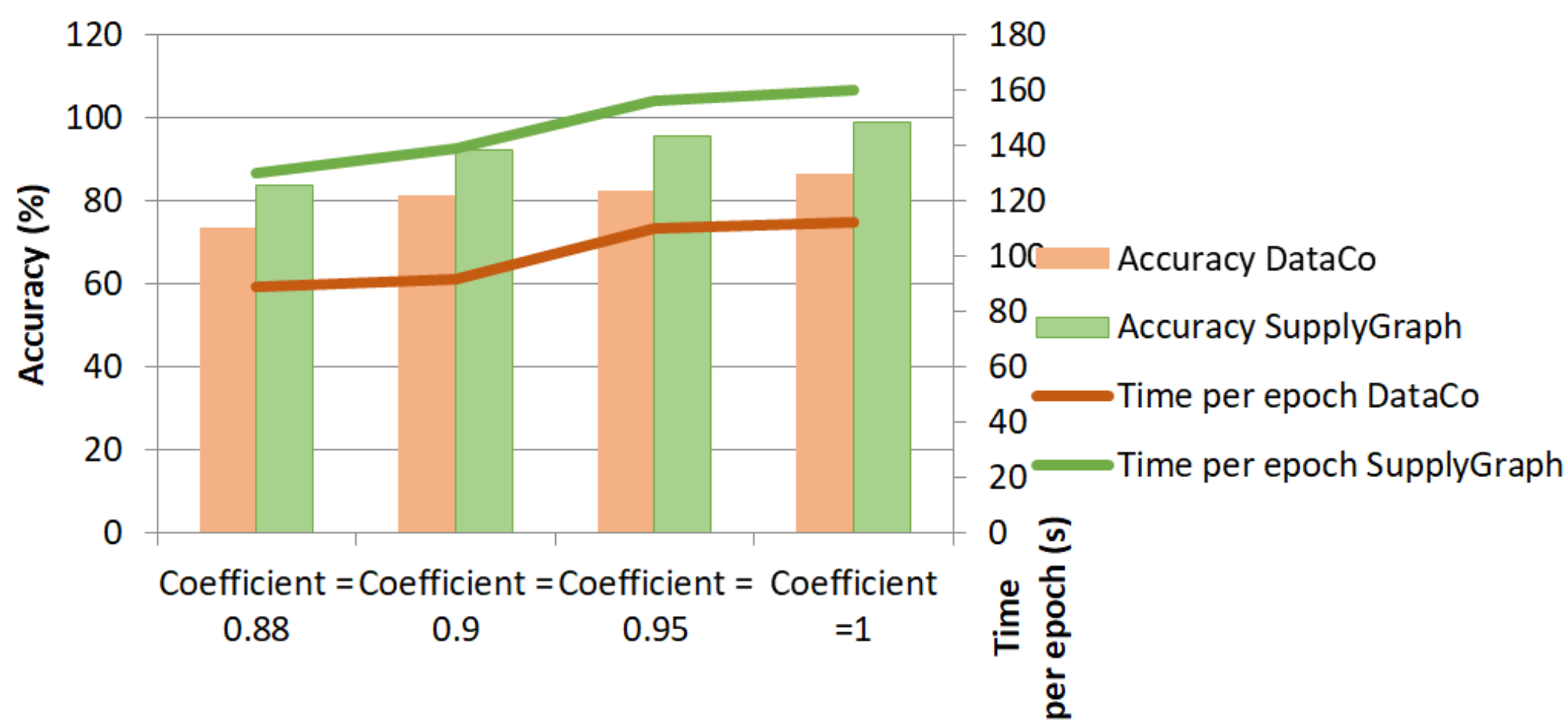

Figure 18. F1-score and Time of Training per Epoch with different threshold level for graph construction.

Figure 18 is the comparison outcome of different altering coefficient $\alpha$ of the ensemble loss function in algorithm 1. This column chart outlines that adjusting the coefficient equal to 0.9 will optimize the accuracy and it is the effective one taking into consideration the converging time.

Figure 19 and 20 are the confusion matrixes for edge classification of the SupplyGraph considering the proposed Ch-EGN. The left matrix in Figure 19 considers the delivery to distributor for nodes and the factory issue has been considered for calculating the right confusion matrix in this figure. Figure 20 considers the sales order for the left one and the time-series production for the right confusion matrix in this figure. All of them confirms the efficiency of the proposed method.

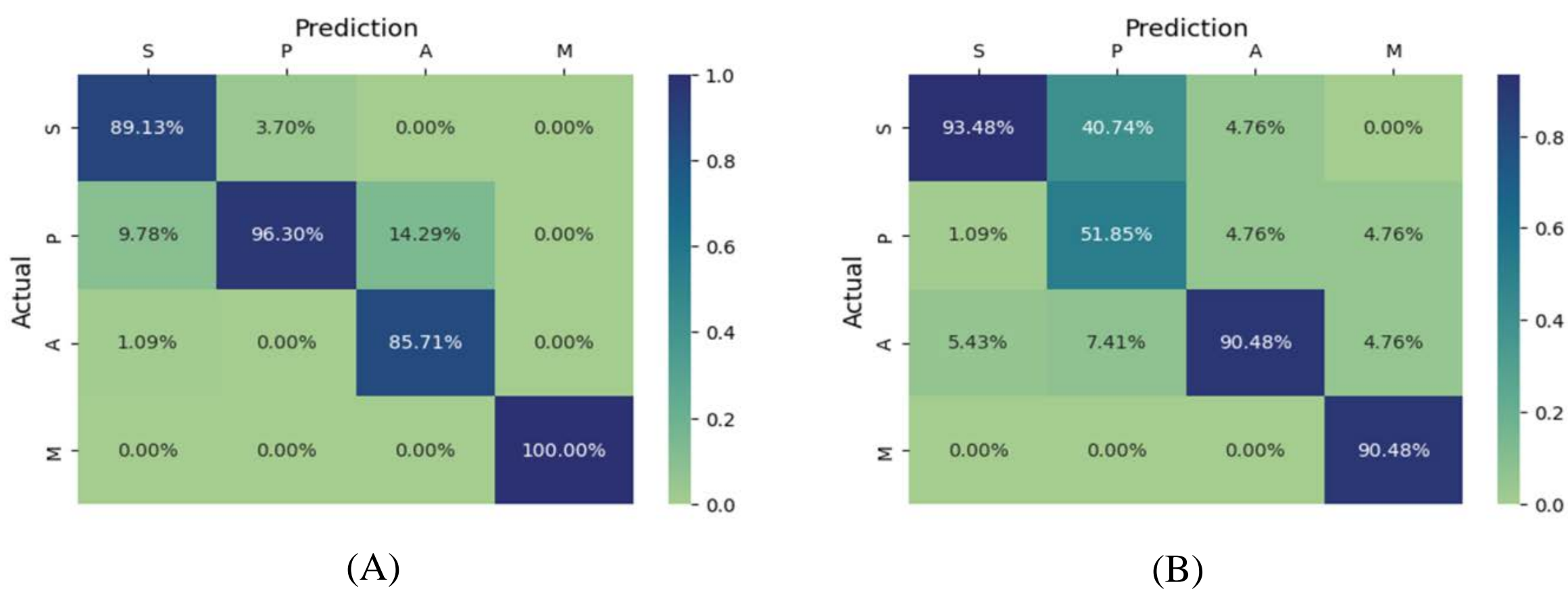

Figure19. The confusion matrix for edge classification for product group connections (A)delivery to distributor, (B)factory issue.

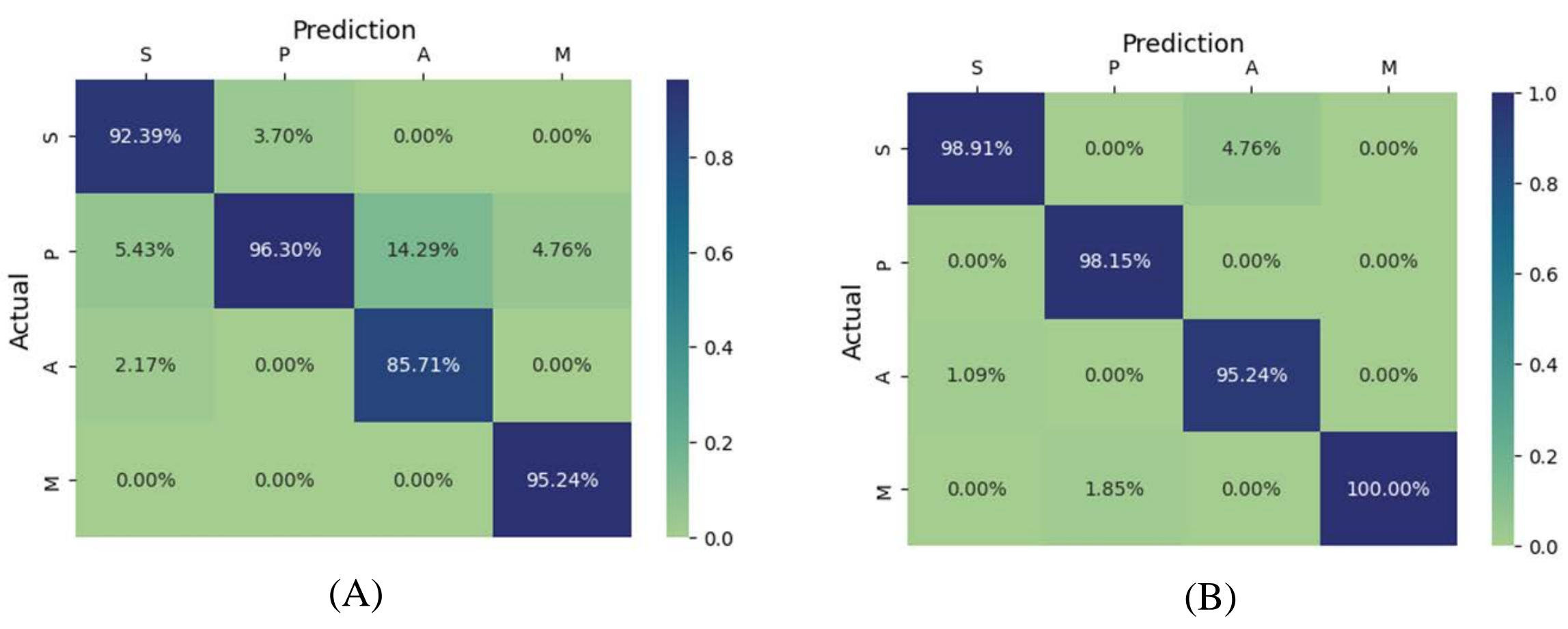

Figure 20. The confusion matrix for edge classification for product group connections (A)sales order, (B)production.

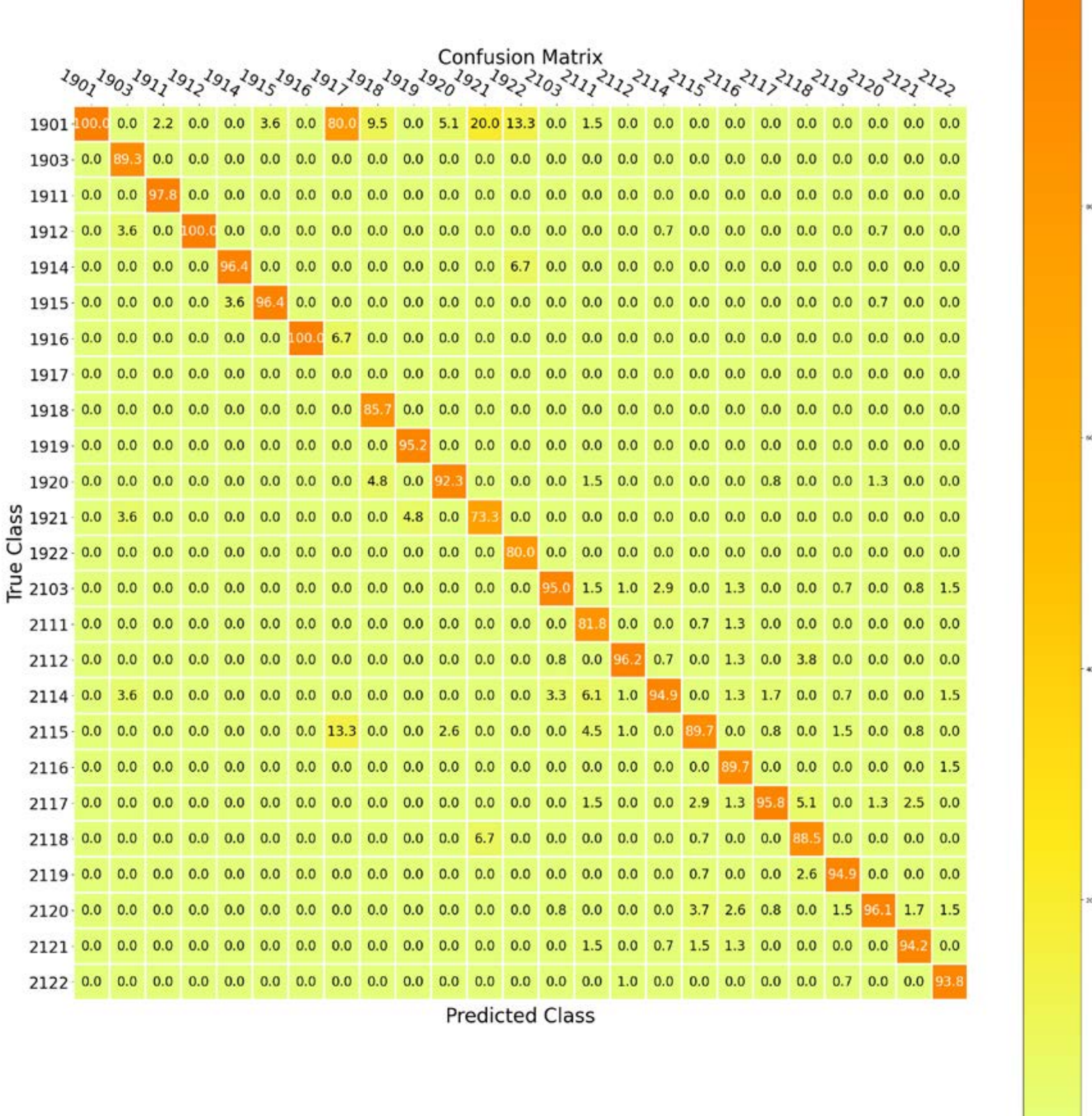

Figure 21. The confusion matrix for edge connections in terms of plant labels.

Figure 21 showcases the confusion matrix for edge classification of the SupplyGraph dataset considering the edge connections for similar plant locations for different types of products.

## 6. Conclusion

In this paper, a novel deep geometric architecture is proposed to solve the problem of prediction of delivery status for risk management in a supply chain. In addition, it is a deep model proposed to strengthening the sustainability of a supply chain. The proposed model architecture is used for testing the sustainability of the SupplyGraph database and it is used for risk management in DataCo dataset.

The main challenges in this paper are removed feature extraction phase and considering the connectivity between nodes to extract the hidden states of supply chain characteristic vectors. The proposed ensemble geometric deep network is a novel approach for node and edge classification. It facilitates the risk management in a supply chain along with strengthening the sustainability of a supply chain. The efficiency of the proposed method is delineated via the exploratory outcomes on the DataCo and the SupplyGraph datasets. Comparing the results with other novel state-of-the-art methods confirms that the Ch-EGN receives higher classification accuracy for node and edge classification with fewer numbers of iterations.